\documentclass{article} 
\usepackage{iclr2027_conference,times}

\usepackage{amsmath,amsfonts,bm}

\def\eqref#1{equation~\ref{#1}}

\def\1{\bm{1}}

\DeclareMathAlphabet{\mathsfit}{\encodingdefault}{\sfdefault}{m}{sl}
\SetMathAlphabet{\mathsfit}{bold}{\encodingdefault}{\sfdefault}{bx}{n}

\usepackage{hyperref}
\usepackage{url}
\usepackage{xparse}
\usepackage{booktabs}
\usepackage{siunitx}
\usepackage{tabularx}
\usepackage{multirow}
\usepackage{graphicx}

\title{LLMs learn different forms of metacognition when trained to predict their own accuracy}

\author{Nicolas Yax \\\small{LNC2, INSERM, Paris, France} \\\small{DEC, ENS, PSL, Paris, France} \\\small{Flowers AI \& CogSci Lab}\\\small{Centre Inria de l'Université de Bordeaux France} \\\small{nicolas.yax@ens.psl.eu} \\\And Stefano Palminteri$^*$\\\small{LNC2, INSERM, Paris, France} \\\small{DEC, ENS, PSL, Paris, France} \\\And Pierre-Yves Oudeyer$^*$ \\ \small{Flowers AI \& CogSci Lab }\\
\small{Centre Inria de l'Université de Bordeaux France} \\\\ $*$ equal contribution \\}

\newcommand{\spre}{s^{\mathrm{pre}}}
\newcommand{\squest}[1]{s^{\mathrm{q}}_{#1}}
\NewDocumentCommand{\sopt}{oo}{%
  s^{\mathrm{opt}}\IfNoValueF{#1}{_{#1\IfNoValueF{#2}{,#2}}}%
}
\newcommand{\sans}[1]{s^{\mathrm{ans}}_{#1}}
\newcommand{\cat}{\oplus}

\iclrfinalcopy 
\begin{document}

\maketitle

\begin{abstract}
Large language models are trained to always produce an answer, regardless of whether they possess the relevant knowledge, which leads them to fabricate facts. Prior work has shown that LLMs' confidence estimates correspond poorly to their actual performance, and that fine-tuning can substantially improve them. However, what models actually learn during such training remains poorly understood. We investigate how LLMs acquire metacognitive monitoring, the ability to know what one knows, by training 10 open-weight LLMs to predict their own accuracy on factual multiple-choice questions before answering them. We find that trained confidence reflects two distinct signals. While on questions close to the training data, it tracks the model's true accuracy, in other domains, it instead tracks output consistency: the concentration of the model's answer distribution. Output consistency tracking emerges early in training and generalizes across datasets, whereas accuracy tracking develops later and remains local to the training distribution. These results suggest that calibration training may not teach models to generally detect errors they commit confidently, and they raise broader questions about the nature of metacognition in artificial systems.
\end{abstract}

\section{Introduction}
Large language models are increasingly integrated into work and daily life, with capabilities spanning coding, mathematical reasoning, creative writing, and information synthesis \citep{griot25}. However, they struggle to recognize the boundaries of their knowledge and often exhibit overconfidence \citep{sun2025largelanguagemodelsoverconfident}. This is partly a consequence of training and evaluation practices that reward always producing an answer, which incentivizes hallucination when the model lacks the relevant information \citep{kalai2025languagemodelshallucinate}. As users increasingly trust model outputs \citep{colombatto25,steyvers2025large}, the absence of a reliable signal indicating when a model is likely to be wrong poses substantial risks \citep{weidinger2021ethicalsocialrisksharm}.

A promising way to address this limitation is to equip LLMs with metacognitive monitoring, the ability to assess their own knowledge states. Prior work has pursued this in several ways. Prompting-based methods elicit verbalized confidence estimates \citep{tian2023justaskcalibrationstrategies,xiong2024llmsexpressuncertaintyempirical}. Sampling-based methods instead estimate uncertainty from the output consistency of a model's answers across samples \citep{kuhn2023semanticuncertaintylinguisticinvariances,manakul2023selfcheckgptzeroresourceblackboxhallucination}. Finally, fine-tuning and probing approaches train models, or lightweight heads on their internal representations, to predict their own correctness \citep{kadavath2022languagemodelsmostlyknow,kapoor2024largelanguagemodelstaught}. These generally yield better confidence estimates than prompting alone.

While prior work has achieved impressive calibration performance \citep{wang2025objectivefinetuningllmsprior,kapoor-etal-2024-calibration}, the research community has focused primarily on achieving high metrics on complex tasks, with little investigation into the underlying metacognitive processes. The mechanisms by which models assess their knowledge remain poorly understood—a critical gap given the importance of such skills in everyday user interactions. Understanding whether models truly assess their knowledge or merely track superficial signals has important implications for deployment: statistical shortcuts may fail under distribution shifts, in adversarial contexts, or when models need to reliably abstain from answering, even if they achieve high performance on standard benchmarks.

In this work, we investigate what LLMs learn when trained to estimate their own performance. To isolate metacognitive monitoring from other sources of error, such as faulty reasoning or planning, we focus on factual multiple-choice questions, which a model either knows the answer to or does not. Following \citet{kadavath2022languagemodelsmostlyknow} we adopt a prospective setting in which the model estimates its accuracy on a question before answering it. This way, confidence reflects whether the model knows the answer rather than whether it can verify an answer it has already given. We fine-tune 10 open-weight LLMs, together with a linear probe on their hidden states, to predict their own accuracy on questions from science, medicine, and arithmetic datasets. Because train and test sets are split by topic, we can evaluate how learned confidence transfers to unseen domains. Beyond standard correlation metrics, we examine how trained confidence relates to other properties of the model's output distribution, how its behavior changes with semantic distance from the training data, and how it emerges over the course of training.

\paragraph{Contributions}
Consistent with prior work, we find that verbalized confidence is essentially uncorrelated with true accuracy before training, and that fine-tuning substantially improves it across all 10 models and 5 datasets. However, we found that trained confidence reflects two distinct signals: it tracks true accuracy on questions close to the training data, but output consistency, i.e., the concentration of the answer distribution, on more distant ones. This distinction matters because the two can differ sharply, for instance, when a model reliably selects a wrong answer. Finally, we find that output consistency tracking emerges early in training, whereas accuracy tracking develops later and remains confined to the training split and similar questions.

\begin{figure}[t]
    \centering
    \includegraphics[width=1.0\linewidth]{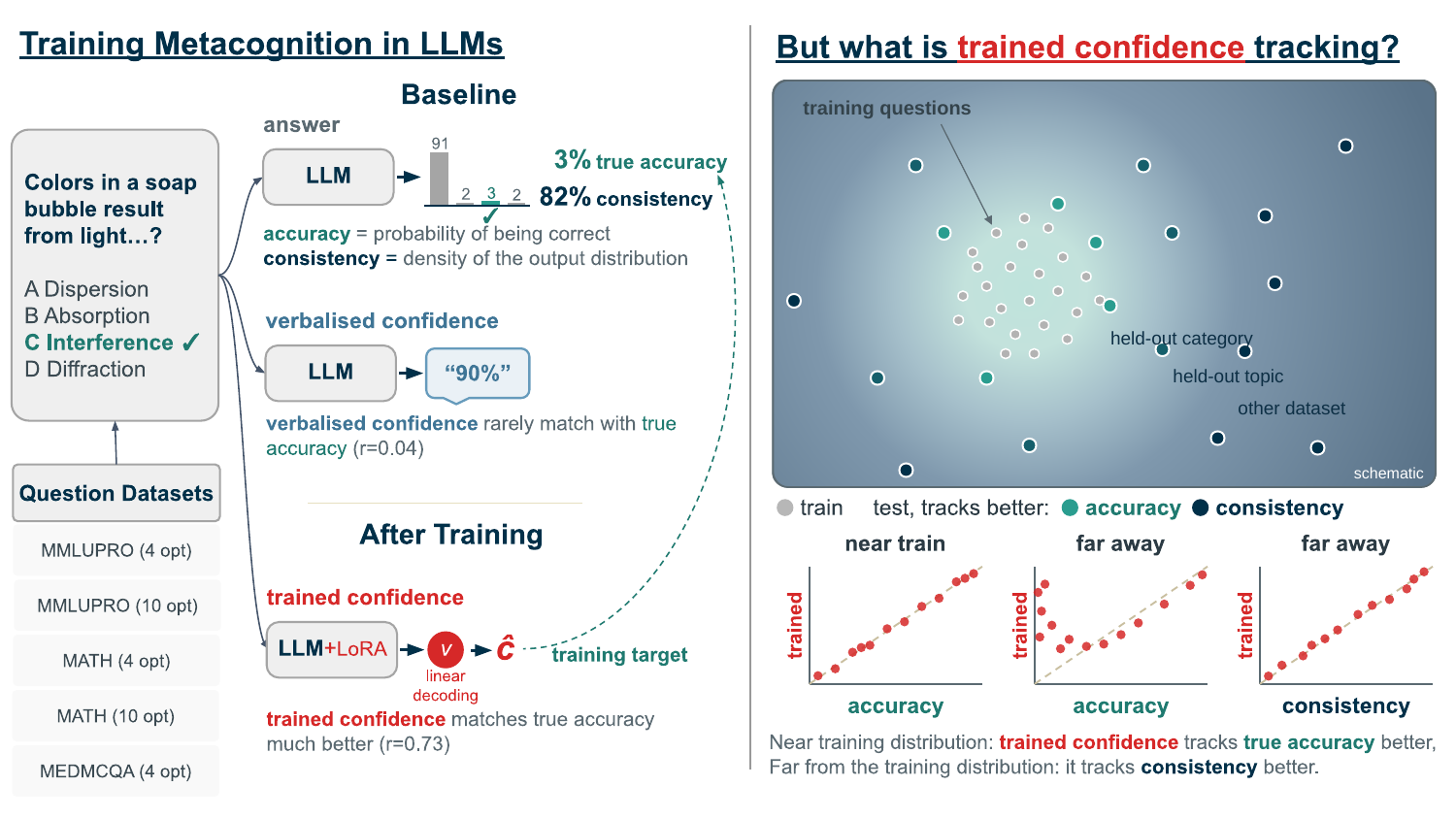}
    \caption{\textbf{Trained confidence tracks accuracy locally and consistency globally} Left side: experimental setup — models answer factual multiple-choice questions from five dataset variants. Accuracy is the probability assigned to the correct option, and output consistency is the normalized concentration of the answer distribution; the two can differ sharply, e.g., when a model reliably selects a wrong answer. Before training, verbalized confidence is essentially uncorrelated with true accuracy ($r = 0.04$). After fine-tuning with LoRA and a linear probe, trained confidence predicts accuracy much better ($r = 0.73$). Right side: schematics of the results — trained confidence tracks true accuracy on questions near the training distribution but output consistency on more distant ones, such as held-out topics or other datasets.}
    \label{fig:main}
\end{figure}

\section{Methods}
As outlined in the introduction, we aim to isolate metacognitive monitoring from other processes. We therefore use multiple-choice questions that models answer directly, without chain-of-thought, a task requiring pure knowledge recall rather than classical reasoning tasks on which multiple metacognitive processes would need to be aggregated to guess the model's own performance. The multiple-choice format also allows us to compute a model's accuracy and the concentration of its answer distribution exactly from the probabilities it assigns to each option. Confidence is elicited prospectively, before the model answers, so that it reflects whether the model knows the answer rather than whether it can verify an answer it has already produced.

\subsection{Datasets}
To evaluate metacognitive performances on factual knowledge we used 3 distinct datasets: \textsc{MMLU-PRO}\citep{wang2024mmluprorobustchallengingmultitask} which includes around 10k  questions on various scientific topics, \textsc{MedMCQA} \citep{pal2022medmcqalargescalemultisubject}, which contains 100k questions from multiple medical domains; and a custom MATH dataset (see Appendix \ref{app:math_dataset}) which includes 100k calculation questions. These datasets include multiple choice questions, and most can be answered without reasoning by most modern AI models.

Let's introduce some notations: all these datasets $\mathcal{D}$ contain questions $\squest{}$ with either 4 or 10 different options depending on the dataset $(\sopt[j])_{j\le M_{\mathcal{D}}}$ (their concatenation is written $\sopt[] = \sopt[0]\sopt[1]\dots\sopt[M_{\mathcal{D}}-1]$) and a single correct answer among these options $\sans{}$. A dataset is written as a list of items $\mathcal{I}$: $\mathcal{D} = (\mathcal{I}_i)_{i\le N_\mathcal{D}} =(\squest{i},(\sopt[i][j])_{j\le M_\mathcal{D}},\sans{i})_{i\le N_\mathcal{D}}$. 

We produced variations of these datasets by making the number of options vary: \textsc{MMLU-PRO}$_{4}$ with 4 options, \textsc{MMLU-PRO}$_{10}$ with 10 options (original), \textsc{MedMCQA}$_4$ (original), \textsc{MATH}$_4$ (custom) and \textsc{MATH}$_{10}$ (custom). Additionally, we removed items longer than 500 tokens (using Mistral-7B tokenizer) for hardware requirements in order to run all experiments on a single H100 80GB (this resulted in around 5\% of the original items removed in some datasets).

In all datasets, items are sorted into various categories. As such, each dataset variant is split into a train set and a test set such that
(i) the accuracy of models remains similar on both the train and test sets; (ii) the train set includes around 90\% of the questions (and the test set includes around 10\%); (iii) for each category of question, (usually the topic of the question, such as Math, Physics, ...), all questions in this category are either in the train set or the test set (i.e. the split is a partitioning of the categories rather than individual questions).

More details about the dataset splitting method are in Appendix~\ref{app:dataset_splitting}. This ensures that the train and test sets are thematically distinct while retaining comparable metrics in accuracy to avoid statistical biases when comparing metacognitive performances between the train and test sets. The categories split are shown in Table~\ref{tab:category_splits} in Appendix~\ref{app:dataset_splitting} and accuracy statistics are reported in Table~\ref{tab:model_performances} in Appendix~\ref{app:dataset_splitting}. In practice, the categories composing \textsc{MATH} datasets are closer than \textsc{MMLU-PRO} and \textsc{MEDMCQA}, as they refer to subcategories of arithmetic computations, while the latter refer to completely different topics.

\subsection{Models}
We evaluated and finetuned 10 different LLMs: Qwen2.5-7B, Qwen2.5-7B-Instruct \citep{qwen2025qwen25technicalreport}, Qwen3.5-9B-Base, Qwen3.5-9B \citep{yang2025qwen3technicalreport}, Llama-2-7b-chat \citep{touvron2023llama2openfoundation}, Llama-3.2-3B-Instruct \citep{grattafiori2024llama3herdmodels}, phi-4 \citep{abdin2024phi4technicalreport}, Mistral-7B-Instruct-v0.3 \citep{jiang2023mistral7b}, Ministral-3-3B-Instruct-2512 and Ministral-3-8B-Instruct-2512 \citep{liu2026ministral3}. These were chosen for covering a wide range of recent development in LLMs, including base and instruct tuned models from various companies and training methods. All models used in this paper are quantized in 4-bits using unsloth \citep{unsloth} and were chosen for their capability to fit on a single H100 80GB GPU. Thus this study does not include very large models for compute limitation reasons.

\subsection{Evaluation procedures}
\label{sec:evaluation_procedures}
\paragraph{Accuracy evaluation}
Models are evaluated on each dataset split by using a question/answer prompt (see Appendix \ref{app:question_formatting} for formatting details) and evaluating the probability to generate each option. The probability associated with the correct option is the accuracy of the model on the question.
\begin{equation}
    Acc(\mathcal{M}_\theta,\mathcal{I}) = p_\theta(\sans{}|\squest{}\sopt[])
\end{equation}
with $\mathcal{M}_\theta$ the model, $\mathcal{I}$ the item composed of a question $\squest{}$, a list of options $(\sopt[j])_{j\in M_\mathcal{D}}$ and $p_\theta(\sans{}|\squest{}\sopt[])$ the generative probability function of model $\mathcal{M}_\theta$ for generating $\sans{}$ after prompt $\squest{}\sopt[]$ (concatenation of the question with the options).

The accuracy of a model on dataset split $\mathcal{D}$ is the average accuracy on the dataset split and is evaluated for both the train and the test splits:
\begin{equation}
    Acc(\mathcal{M}_\theta,\mathcal{D})=\frac{1}{N_\mathcal{D}}\sum_{i\le N_\mathcal{D}}Acc(\mathcal{M}_\theta,\mathcal{I}_i)
\end{equation}

\paragraph{Output consistency evaluation}
Aside from accuracy, the output consistency of the model on each question was also evaluated. Output consistency is a measurement of how often the LLM would give the same answer to the same question and is mathematically defined as
\begin{equation}
    Con(\mathcal{M}_\theta,\mathcal{I}) = 1-\frac{H(p_\theta(.|\squest{}\sopt[]))}{\log(M_\mathcal{D})}
\end{equation}
with $M_\mathcal{D}$ the number of options for dataset $\mathcal{D}$ from which item $\mathcal{I}$ is extracted, $H$ the entropy function and $p_\theta(.|\squest{}\sopt[])$ the probability distribution of sampling one of the possible options in $(\sopt[i])_{i\le M_\mathcal{D}}$.

In other words, the output consistency is the entropy of the output distribution (only considering the possible options) scaled in [0,1] such that an output consistency of 0 corresponds to a uniform distribution and an output consistency of 1 represents an output distribution associating probability 1 to one of the options. Output consistency evaluation is performed on both train and test splits.

\paragraph{Metacognitive evaluation}
\label{par:metacognitive_evaluation}
Lastly, metacognitive performance is measured using two different methods. Before training, it is evaluated with a verbal statement from the LLM using an additional pre-prompt $\spre{}$ before the question to ask the LLM to give its accuracy estimate on the given question instead of answering it (see Appendix \ref{app:question_formatting} for prompting details). 
\begin{equation}
    Met_{verb}(\mathcal{M}_\theta,\mathcal{I}) = \text{parse\_number}\circ\text{greedy}(\mathcal{M}_\theta,\spre{}\squest{}\sopt[],5)
\end{equation}
with $\text{greedy}(\mathcal{M}_\theta,t,n)$ the function that autoregressively generates $n$ tokens after context $t$ using LLM $\mathcal{M}_\theta$ and $\text{parse\_number}$ the function that parses the first number in a text (transforms from string to integer/float).

After training, the procedure involves reading from the trained probe $\mathbf{v}$ at the end of the transformer stack (see Section \ref{sec:training}). The output of the probe is passed through a sigmoid function to scale it between 0 and 1.
\begin{equation}
    Met_{prob}(\mathcal{M}_\theta,\mathcal{I}) = \sigma(\mathbf{v}\cdot \mathbf{t}_\theta^{end}(\spre{}\squest{}\sopt[]))
\end{equation}
with $\mathbf{t}_\theta^{end}(\spre{}\squest{}\sopt[])$ the function that returns the output at the end of the transformer stack of LLM $\mathcal{M}_\theta$ after input $\spre{}\squest{}\sopt[]$, $\sigma$ the sigmoid function and $\mathbf{v}$ a vector learnt during training. Another training setup puts the probe at the middle of the transformer stack instead of the end. In that case it is $\mathbf{t}_\theta^{mid}(\spre{}\squest{}\sopt[])$ that is used, namely the latent activation at the middle of the transformer stack.

\subsection{Training Procedures}
\label{sec:training}
The training setup consists in adding LoRA adapters in the original quantized LLM used to compute the accuracy, output consistency and verbalised metacognition (QLoRA \citep{dettmers2023qloraefficientfinetuningquantized}) and adding a linear regression (called probe) from the output activations to a single scalar at the end of the transformer stack. This setup is inspired from \citet{kadavath2022languagemodelsmostlyknow} who shown that adding a probe yields better results than training token generation and later from \citet{gaven2025magellanmetacognitivepredictionslearning} that implemented it using QLoRA. 

Models were trained on the train split of each dataset variant 5 times with an MSE loss to predict their true accuracy:
\begin{equation}
    Loss(\phi) = \mathbb{E}_{\mathcal{I}\in \mathcal{D}}\left[(Met_{prob}(\mathcal{M}_{\phi},\mathcal{I})-Acc(\mathcal{M}_\theta,\mathcal{I}))^2\right]^{\frac{1}{2}}
\end{equation}
This loss on $\phi$ was minimized on the train set for a single epoch for variants of both \textsc{MedMCQA} and \textsc{MATH} datasets (100k items in the train split is 100k steps) and for 8 epochs for \textsc{MMLU-PRO} (10k items in train split $\times 8=80$k steps) using QLoRA to approach the same quantity of training steps than \textsc{MedMCQA} and \textsc{MATH}. Hyperparameters are detailed in Appendix \ref{app:hyperparameters}.

Training metrics and training curves are reported in Appendix \ref{app:training_curves}. In this experiment, we train metacognition alone: the additional LoRA adaptors can only be used to predict metacognition and make the network unable to answer questions anymore.

\subsection{Analyses}

\subsubsection{Correlation metrics}
We analyzed the data using Pearson correlations between verbalized confidence, true accuracy, and output consistency. The choice of metrics was motivated by the fact that our data are continuous and that we aim to quantify how much of a linear relationship exists between two variables. More precisely, we used differences in Pearson correlations to quantify how much a trained confidence linearly reflects true accuracy or output consistency:
\begin{equation}
\begin{split}
 \Delta r(\mathcal{M}_\theta,\mathcal{D}) =&r\left[(Acc(\mathcal{M}_\theta,\mathcal{I}))_{\mathcal{I}\in\mathcal{D}},(Met_{prob}(\mathcal{M}_\theta,\mathcal{I}))_{\mathcal{I}\in\mathcal{D}}\right]\\ &- r\left[(Con(\mathcal{M}_\theta,\mathcal{I}))_{\mathcal{I}\in\mathcal{D}},(Met_{prob}(\mathcal{M}_\theta,\mathcal{I}))_{\mathcal{I}\in\mathcal{D}}\right]
\end{split}
\end{equation}
for some (trained) model $\mathcal{M}_\theta$ tested on dataset $\mathcal{D} = (\mathcal{I}_i)_{i\le N_\mathcal{D}}$.

\subsubsection{Error metrics}
While correlation can be very useful for studying how well a LLM is predicting its accuracy on a dataset, it cannot measure whether the answer to an individual question reflects true accuracy or rather output consistency. For this specific situation, we used the differences in absolute errors for some (trained) model $\mathcal{M}_\theta$ tested on question $\mathcal{I}$:
\begin{equation}
 \Delta e(\mathcal{M}_\theta,\mathcal{I}) = \lVert Acc(\mathcal{M}_\theta,\mathcal{I})-Met_{prob}(\mathcal{M}_\theta,\mathcal{I})\rVert - \lVert Con(\mathcal{M}_\theta,\mathcal{I})-Met_{prob}(\mathcal{M}_\theta,\mathcal{I})\rVert
\end{equation}
To recap, the correlation is used to compute the performance on a set of questions (usually a dataset), while the error is used to compute the performance on a single question. Indeed, in this study, we investigate whether trained confidence matches a linear or U-shape pattern with true accuracy or consistency. Therefore correlation is a better suited tool to study this dependency at the dataset level than error (Brier score) so we only keep it for individual questions and more specific experiment.

\subsubsection{Embeddings}
\label{embeddings}
We embed each question using the Qwen3-Embedding-0.6B model \citep{zhang2025qwen3embeddingadvancingtext}, which maps text to a 1024-dimensional vector such that semantically similar texts have nearby representations. We use these embeddings to quantify how similar questions are, both within and across datasets, and compute them on the training and test splits of each dataset. Each item $\mathcal{I}$ is embedded together with its options ($\squest{}\sopt[]$), giving an embedding $e_\mathcal{I}$; the embeddings of the same question in the 4- and 10-option variants of a dataset are therefore similar but not identical. The distance between two items is their cosine distance, consistent with the similarity measure used to train the embedding model, and we compute the distance from a test item $\mathcal{I}$ to a training dataset $\mathcal{D}$ as the average distance to its $k = 100$ nearest training items where $\mathcal{N}_k(\mathcal{I}, \mathcal{D}) \subset \mathcal{D}$ is the set of the $k$ training items closest to $\mathcal{I}$:
\begin{equation}
\label{eq:dist_set}
d_k(\mathcal{I}, \mathcal{D}) = \frac{1}{k} \sum_{\mathcal{I}' \in \mathcal{N}_k(\mathcal{I}, \mathcal{D})}\left( 1 - \frac{e_\mathcal{I} \cdot e_{\mathcal{I}'}}{\lVert e_\mathcal{I} \rVert \, \lVert e_{\mathcal{I}'} \rVert}\right),
\end{equation}

\section{Results}
We first compare confidence estimates before and after training (Section~\ref{sec:perf}). We then analyze which signals trained confidence tracks and how this depends on the similarity between test questions and the training data (Sections~\ref{sec:signals}--\ref{sec:distance}). Finally, we also examine how these signals emerge over the course of training (Appendix~\ref{sec:development}).

\subsection{Fine-tuning substantially improves confidence estimation}
\label{sec:perf}
As a baseline, we elicit verbalized confidence estimates from untrained models. These estimates are essentially uncorrelated with true accuracy on the test splits ($r = 0.04 \pm 0.01$, mean $\pm$ s.e.m.\ across models and datasets). After fine-tuning, confidence estimates correlate strongly with accuracy on test splits ($r = 0.73 \pm 0.03$). This improvement holds for every model and dataset (Appendix~\ref{app:individual_plots} Figure~\ref{fig:trained_perf} for the details). Note that the two conditions differ both in training and in elicitation format (verbal report vs. linear probe): since an untrained probe produces arbitrary outputs, verbalized confidence is the natural baseline before any form of training. This step simply builds the models we are going to investigate in the rest of the paper.

\subsection{Trained confidence tracks true accuracy or output consistency depending on the dataset}
\label{sec:signals}
\begin{figure}[h]
    \centering
    \includegraphics[width=0.8\linewidth]{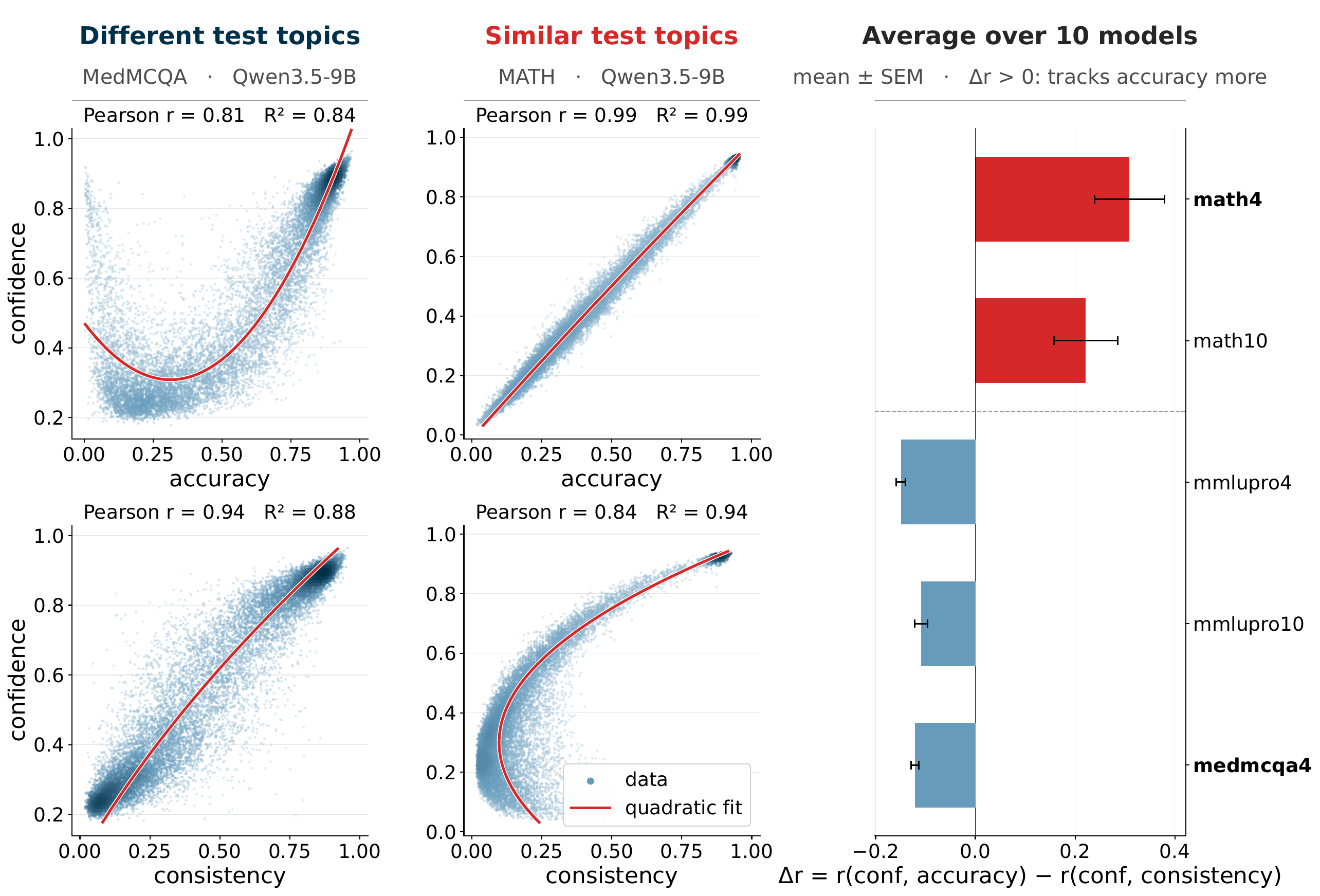}
    \caption{\textbf{Relationship between predicted accuracy, true accuracy and output consistency.} Left panel shows the trained confidence of Qwen3.5-9B against true accuracy (top) and output consistency (bottom) on the test split of \textsc{MedMCQA}$_4$ and \textsc{Math}$_4$. Red lines are quadratic fits ($R^2$ shown above each plot). Right panel represents $\Delta r$ on each test split; positive values (red) indicate that confidence tracks true accuracy more than output consistency, negative values (blue) the reverse. Error bars: s.e.m.\ across the 10 models (each averaged over 5 runs).}
    \label{fig:accuracy_consistency}
\end{figure}

To examine the learned relationship at the level of individual questions, we plot trained confidence against true accuracy. Figure~\ref{fig:accuracy_consistency} (top left) shows this relationship for Qwen3.5-9B trained on the training split of \textsc{MedMCQA}$_4$ and evaluated on its test split, using data from the first training run. The relationship follows a U-shape: confidence closely matches accuracy on questions the model answers correctly with high probability, slightly underestimates accuracy on questions of intermediate accuracy, and strongly overestimates it on many questions with accuracy close to zero, i.e., questions on which the model reliably selects an incorrect answer. Moreover, the model never predicts an accuracy below chance level (25\% in this example), although many questions fall below this threshold because the model consistently selects the same incorrect answer.

In contrast, plotting trained confidence against output consistency for the same model and test split reveals an almost linear relationship (Figure~\ref{fig:accuracy_consistency}, bottom left), with a higher correlation than with true accuracy ($r = 0.94$ vs.\ $r = 0.81$). This suggests that the model may use its output consistency as a proxy for accuracy. Indeed, a model that performs well on a task achieves high accuracy by consistently selecting the correct answer. Consequently, the more competent the model, the better its output consistency on a question predicts whether it will answer correctly.

This pattern reverses on \textsc{Math}$_4$ (Figure~\ref{fig:accuracy_consistency}, right column of the left panel), where Qwen3.5-9B is trained on the training split and evaluated on the test split of \textsc{Math}$_4$, again using data from the first run on this training set. Here, confidence tracks true accuracy almost perfectly ($r = 0.99$), whereas its relationship with output consistency is non-linear ($r = 0.84$) and follows a transposed U-shape.

These observations generally hold across models. On \textsc{MMLU-Pro}$_4$, \textsc{MMLU-Pro}$_{10}$, and \textsc{MedMCQA}$_4$, most models show a U-shaped relationship between trained confidence and true accuracy (more details about this in Appendix~\ref{app:individual_plots}), and a nearly linear relationship between trained confidence and output consistency. To quantify this difference, Figure~\ref{fig:accuracy_consistency} (right panel) reports $\Delta r$, the difference between the correlation of trained confidence with true accuracy and trained confidence correlation with output consistency, computed on the test split of each dataset. On these three datasets (blue), $\Delta r$ is negative ($\Delta r = -0.12 \pm 0.01$, mean $\pm$ s.e.m.\ across models), indicating that trained confidence is more linearly related to output consistency than to true accuracy. Conversely, on the \textsc{Math} datasets (red), trained confidence is almost linearly related to true accuracy, while its relationship with output consistency follows a U-shape ($\Delta r = 0.27 \pm 0.06$).

\subsection{Output consistency tracking transfers across datasets}
\label{sec:generalization}

\begin{figure}[h]
    \centering
    \includegraphics[width=1.0\linewidth]{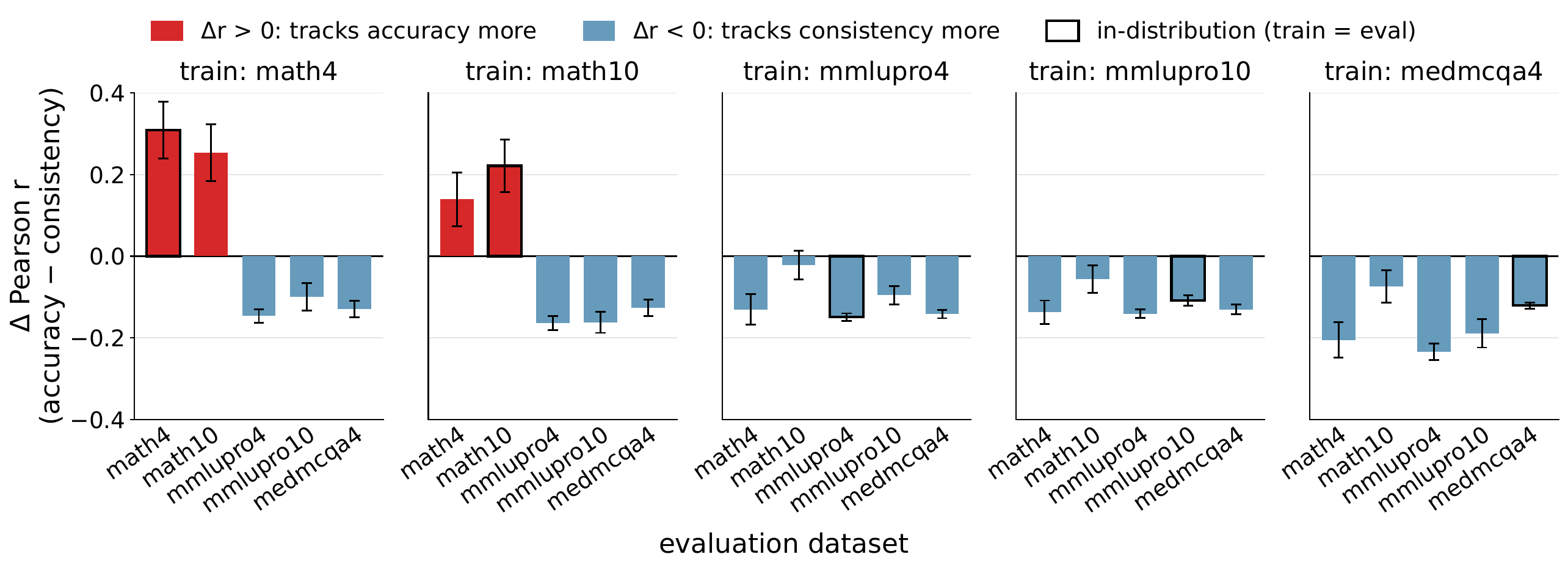}
    \caption{\textbf{Cross-dataset evaluation of trained confidence.} Each panel shows models trained on one dataset and evaluated on the test split of all five datasets. Bars represent $\Delta r$; positive values (red) indicate that confidence tracks true accuracy more than output consistency, negative values (blue) the reverse. In-distribution evaluations (tested on the test split from the training dataset - thematically distinct but closer with respect to the other datasets) are highlighted. Error bars: s.e.m.\ across the 10 models (each averaged over 5 runs).}
    \label{fig:cross_perf}
\end{figure}

We hypothesize that this difference reflects the distance between training and test questions. Although all splits are category-disjoint, the categories of the \textsc{Math} datasets are much closer to each other than those of \textsc{MMLU-Pro} and \textsc{MedMCQA}: the \textsc{Math} test category (easy$*$) lies between training categories (easy$-$, easy$+$, medium$*$ and hard$*$), whereas the test categories of the other datasets (e.g., philosophy or ophthalmology) are distinct topics. To test this hypothesis, we evaluate each model in a 5x5 design: each model (trained on train split of dataset X) is tested on all datasets test split (Figure~\ref{fig:cross_perf}). This cross evaluation further increases the distance between training and test questions. If our hypothesis holds, models should track output consistency on all datasets other than their training dataset, with the exception of the two \textsc{Math} and \textsc{MMLU-PRO} variants, which share the same questions and categories.

Consistent with this prediction, models trained on \textsc{Math} datasets linearly track true accuracy when evaluated on either \textsc{Math} variant ($\Delta r = 0.23 \pm 0.07$), but switch to tracking output consistency when evaluated on \textsc{MMLU-Pro} or \textsc{MedMCQA} ($\Delta r = -0.14 \pm 0.02$). Conversely, models trained on \textsc{MMLU-Pro} or \textsc{MedMCQA} keep tracking output consistency across all evaluation datasets ($\Delta r = -0.13 \pm 0.02$). Accuracy tracking on \textsc{Math} may therefore not be a property of training on the math questions themselves, but arises only when models are trained on nearby questions akin to the model being familiar with its performance on math but not on more general scientific topics.

\subsection{Distance to the training data determines which signal dominates}
\label{sec:distance}

The previous results suggest that trained confidence reflects two signals: true accuracy, which is linearly tracked only for questions close to the training data, and output consistency, which is tracked in datasets far from the training distribution. Nonetheless, this hypothesis has only been tested at the level of the full dataset, namely whether predicted confidence on the full test set reflects generally more true accuracy or output consistency. However, a test set could include some questions that are predicted to be closer to true accuracy and others that better reflect output consistency.

To examine this relationship even closer at the level of individual questions within each dataset, we plot one point per training dataset (5 datasets) per test question (dataset test splits = 38581 test questions) showing $\Delta e$, the difference between the true accuracy and output consistency prediction errors, computed after averaging accuracy, output consistency, and confidence across all models trained on this training set (10 base LLMs x 5 runs). This experiment makes it possible to explore this relationship beyond pure linear or U-shape tracking at the dataset scale and is therefore complementary to the correlation studies conducted so far.



\begin{figure}[h]
    \centering
    \includegraphics[width=0.8\linewidth]{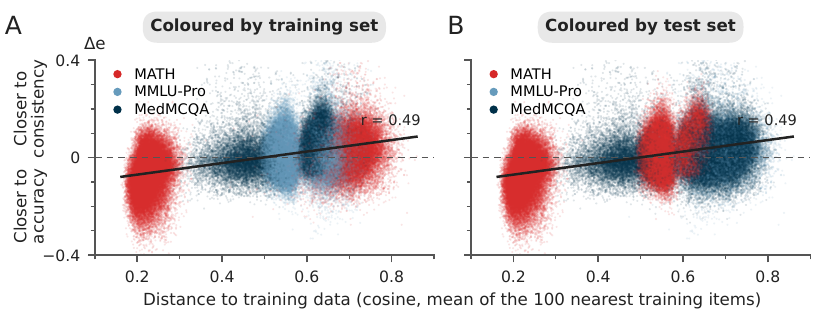}
\caption{\textbf{Distance to the training data and the signal tracked by trained confidence.}
Each point is one test question for one training dataset; its y-value is $\Delta e$
computed from accuracy, consistency and confidence averaged over 10 models $\times$ 5 training runs. Negative values mean confidence is closer to true accuracy,
positive values closer to output consistency. Points are colored by training dataset (A) or
test dataset (B). The x-axis is the mean cosine distance to the 100 nearest training questions.
Black line: linear fit ($r = 0.49$); dashed: $\Delta e = 0$; 0.13\% points beyond
$\pm 0.4$ not shown.}
    \label{fig:distance}
\end{figure}

Figure~\ref{fig:distance} shows that $\Delta e$ increases steadily with distance from the training data for all models and training datasets. Test questions that are similar to at least one training question yield confidence estimates that track true accuracy, whereas questions dissimilar to all training questions yield estimates that track output consistency, regardless of the training dataset.

\section{Discussion}
We investigated what LLMs learn when trained to predict their own accuracy. Although models are explicitly trained to predict true accuracy, their confidence reflects two distinct signals. On questions close to the training data, confidence tracks true accuracy, whereas on more distant questions it tracks output consistency better. We also show that output consistency tracking emerges early in training (see Appendix~\ref{sec:development}) and transfers across datasets, while accuracy tracking develops later and remains confined to questions similar to those seen during training. 

Another way to understand these processes is to see output consistency tracking as a form of first-person metacognition: the model reads out a property of its own internal state, namely how concentrated its beliefs over the answer options are. Because this signal is available for any question, it generalizes across domains and is learned quickly. True accuracy tracking resembles a form of third-person metacognition: by observing its own performance on training questions, the model learns how well it performs on specific topics, much as an external observer could learn to predict the model's behavior from examples. This form of accuracy tracking can be seen as implicit contamination. The model does not generalize whether it knows the answer to a new question, akin to knowing someone's movie preferences is not necessarily informative about their food tastes, but the model rather learns its own accuracy on questions it has seen and on closely related ones. This is consistent with our results on \textsc{MATH} datasets, whose test split is category-disjoint from the training split yet highly similar to it. The scope of third-person metacognition is therefore bounded both by the model's capacity and by its exposure to its own behavior, and does not easily generalize to other types of questions (\textsc{MMLU-PRO} and \textsc{MedMCQA}). Additionally, Appendix~\ref{app:middle_probe} further shows that this dissociation persists when the probe is placed at the middle of the network rather than at its end. Consistency tracking is preserved, whereas true accuracy tracking largely disappears. First, this shows that consistency tracking does not stem from a leak from the network's output, where consistency is directly encoded in the logits — which are computed from the output of the network. Second, it suggests that output consistency tracking may be the simplest and lightweight form of metacognition of the two.

This picture qualifies the concern raised in the introduction that trained confidence may rely on shortcuts that fail under distribution shift. In our experiments, the signal that fails to transfer is accuracy tracking, whereas output consistency tracking, though less precise, is the one that generalizes.

Why would output consistency be the signal that generalizes? A model has no direct access to the truth, only to its own beliefs. LLMs are trained on large corpora that contain inaccurate information, so a model may hold a false belief as firmly as a true one, and both can produce the same internal signal. From the model's perspective, output consistency with its own beliefs is then the best available estimate of correctness, and errors made with high output consistency may not be detectable from the inside. More generally, no cognitive system is omniscient, and it is unclear whether a finite system can fully represent the limits of its own knowledge: this is similar to a map that would have to represent the world at the same time as a complete drawing of itself - the information capacity may prevent it.\footnote{Informally, consider a system with $n$ bits of capacity, $m$ of which encode knowledge. A complete description of its knowledge limits must be stored in the remaining $n-m$ bits, yet must cover both the $m$ bits of knowledge and the $n-m$ bits of the description itself, i.e., information about all $n$ bits.} Perfect metacognition may therefore be out of reach, and the more relevant question is what the upper bound on self-knowledge is. We speculate that combining a general first-person signal with third-person knowledge of one's performance on familiar domains may approach this bound.

These results raise the question of what metacognitive monitoring actually is, not only in LLMs but also in humans. \citet{caziot2021perceptual} showed that human perceptual confidence reflects self-consistency more closely than true accuracy, suggesting that consistency may be the true basis of metacognition. Our results are consistent with this view. Individuals do not have access to the truth, only to their perception of it, so the best estimate of whether one is correct is whether one's answer is consistent with one's own beliefs. Confidence should therefore reflect self-consistency rather than true accuracy. Our models illustrate this in an artificial system: although they are explicitly trained to predict their true accuracy, the optimal behavior found by the optimizer is a signal that generalizes beyond familiar questions: output consistency. Nonetheless, despite these similarities the comparison has a limitation: these two studies do diverge in the exact definition of consistency: in our study output consistency is the peakiness of the output distribution, while in \citet{caziot2021perceptual}, self-consistency is the probability to answer the same answer again. \citet{kumaran2026reportedconfidencellmstracks} has shown that verbal confidence in LLMs tends to match better with commitment in a commit vs. opt-out task, where commitment could be comparable to self-consistency in \citet{caziot2021perceptual}.


Finally, these results raise the question of how a model arbitrates between the two signals for a given question. Our results suggest that distance to the training data is associated with this balance — a form of implicit contamination vs. generalization — but the underlying mechanisms should be studied further. Answering this question would require understanding how a model assesses whether the best way to predict its performance is through its internal consistency or its knowledge of its observed behavior: a form of meta-metacognition.

Our study has several limitations. We focus on factual multiple-choice questions answered without chain-of-thought, and on models between 3B and 9B parameters quantized to 4 bits showing only a small spectrum of LLM confidence interactions. It only includes a couple of datasets and in the training section confidence is compared before and after training using different elicitation formats (verbal report vs.\ linear probe). Still this does not affect most findings in the paper. Finally, distance to the training data is measured with a single embedding model. Extending these analyses to open-ended generation, reasoning tasks, and larger models is an important direction for future work.

\section*{Acknowledgments}
This work was granted access to the HPC/IA resources of [IDRIS HPE Jean Zay A100] under the allocation 2024- [AD011013693R4] and [IDRIS HPE Jean Zay H100] under the allocation 2024-[A0171011996] and  made by GENCI. SP is supported by the European Research Council under the European Union's Horizon 2020 research and innovation program (ERC) (RaReMem: 101043804), the Agence National de la Recherche (CogFinAgent: ANR-21-CE23-0002-02; RELATIVE: ANR-21-CE37-0008-01; RANGE: ANR-21-CE28-0024-01; ANR-17-EURE-0017), the Alexander Von Humboldt foundation, the EUR Frontiers in Cognition, the Idex PSL and a Google unrestricted gift. The Département d'Etudes Cognitives is funded by the Agence Nationale pour la Recherche (ANR-17-EURE-0017, ANR-10- IDEX-0001-02). This work has received support under the Major Research Program of PSL Research University "PSL-Neuro" launched by PSL Research University and implemented by ANR (ANR-10-IDEX-0001). PYO is supported by ANR AI individual chair ANR-19-CHIA-0004.

\section*{AI use statement}
In this work, we have used AI to help implement methods (Claude Code to improve code performance) but we did not use AI tools for any of these tasks: Generate synthetic data sets, help develop theoretical models or conceptual frameworks, formulate mathematical claims, provide critical ingredients for proving mathematical claims, assist in the writing of proofs, propose or refine hypotheses, design or provide feedback on research  methodology or experiments, assist with translation, clean and reformat dataset, support qualitative and thematic data analysis, interpret results. We used AI for these tasks: help generate some of the figures (code and help draft/adjust figure 1 half by hand and half by Claude Opus), format a couple of references in bibtex, edit and generate parts of code, edit a research paper to improve readability and help format some of the appendices and tables from the code and raw results. We have reviewed all AI-assisted work: we have read and updated most of the generated readability and writing suggestions multiple times by all authors, and we checked and adjusted by hand most of the code proposed by AI. We take responsibility for the final content of this work, including text, claims or artifacts produced with the aid of generative AI.

\section*{Reproducibility statement}
Code will be available here \href{https://github.com/Nicolas-Yax/LLM-MetaCog}{https://github.com/Nicolas-Yax/LLM-MetaCog}.

\bibliography{iclr2026_conference}

@misc{gaven2025magellanmetacognitivepredictionslearning,
      title={MAGELLAN: Metacognitive predictions of learning progress guide autotelic LLM agents in large goal spaces}, 
      author={Loris Gaven and Thomas Carta and Clément Romac and Cédric Colas and Sylvain Lamprier and Olivier Sigaud and Pierre-Yves Oudeyer},
      year={2025},
      eprint={2502.07709},
      archivePrefix={arXiv},
      primaryClass={cs.AI},
      url={https://arxiv.org/abs/2502.07709}, 
}

@misc{wang2024mmluprorobustchallengingmultitask,
      title={MMLU-Pro: A More Robust and Challenging Multi-Task Language Understanding Benchmark}, 
      author={Yubo Wang and Xueguang Ma and Ge Zhang and Yuansheng Ni and Abhranil Chandra and Shiguang Guo and Weiming Ren and Aaran Arulraj and Xuan He and Ziyan Jiang and Tianle Li and Max Ku and Kai Wang and Alex Zhuang and Rongqi Fan and Xiang Yue and Wenhu Chen},
      year={2024},
      eprint={2406.01574},
      archivePrefix={arXiv},
      primaryClass={cs.CL},
      url={https://arxiv.org/abs/2406.01574}, 
}

@misc{pal2022medmcqalargescalemultisubject,
      title={MedMCQA : A Large-scale Multi-Subject Multi-Choice Dataset for Medical domain Question Answering}, 
      author={Ankit Pal and Logesh Kumar Umapathi and Malaikannan Sankarasubbu},
      year={2022},
      eprint={2203.14371},
      archivePrefix={arXiv},
      primaryClass={cs.CL},
      url={https://arxiv.org/abs/2203.14371}, 
}

@misc{kadavath2022languagemodelsmostlyknow,
      title={Language Models (Mostly) Know What They Know}, 
      author={Saurav Kadavath and Tom Conerly and Amanda Askell and Tom Henighan and Dawn Drain and Ethan Perez and Nicholas Schiefer and Zac Hatfield-Dodds and Nova DasSarma and Eli Tran-Johnson and Scott Johnston and Sheer El-Showk and Andy Jones and Nelson Elhage and Tristan Hume and Anna Chen and Yuntao Bai and Sam Bowman and Stanislav Fort and Deep Ganguli and Danny Hernandez and Josh Jacobson and Jackson Kernion and Shauna Kravec and Liane Lovitt and Kamal Ndousse and Catherine Olsson and Sam Ringer and Dario Amodei and Tom Brown and Jack Clark and Nicholas Joseph and Ben Mann and Sam McCandlish and Chris Olah and Jared Kaplan},
      year={2022},
      eprint={2207.05221},
      archivePrefix={arXiv},
      primaryClass={cs.CL},
      url={https://arxiv.org/abs/2207.05221}, 
}

@misc{kapoor2024largelanguagemodelstaught,
      title={Large Language Models Must Be Taught to Know What They Don't Know}, 
      author={Sanyam Kapoor and Nate Gruver and Manley Roberts and Katherine Collins and Arka Pal and Umang Bhatt and Adrian Weller and Samuel Dooley and Micah Goldblum and Andrew Gordon Wilson},
      year={2024},
      eprint={2406.08391},
      archivePrefix={arXiv},
      primaryClass={cs.LG},
      url={https://arxiv.org/abs/2406.08391}, 
}

@misc{dettmers2023qloraefficientfinetuningquantized,
      title={QLoRA: Efficient Finetuning of Quantized LLMs}, 
      author={Tim Dettmers and Artidoro Pagnoni and Ari Holtzman and Luke Zettlemoyer},
      year={2023},
      eprint={2305.14314},
      archivePrefix={arXiv},
      primaryClass={cs.LG},
      url={https://arxiv.org/abs/2305.14314}, 
}

@misc{tian2023justaskcalibrationstrategies,
      title={Just Ask for Calibration: Strategies for Eliciting Calibrated Confidence Scores from Language Models Fine-Tuned with Human Feedback}, 
      author={Katherine Tian and Eric Mitchell and Allan Zhou and Archit Sharma and Rafael Rafailov and Huaxiu Yao and Chelsea Finn and Christopher D. Manning},
      year={2023},
      eprint={2305.14975},
      archivePrefix={arXiv},
      primaryClass={cs.CL},
      url={https://arxiv.org/abs/2305.14975}, 
}

@misc{kalai2025languagemodelshallucinate,
      title={Why Language Models Hallucinate}, 
      author={Adam Tauman Kalai and Ofir Nachum and Santosh S. Vempala and Edwin Zhang},
      year={2025},
      eprint={2509.04664},
      archivePrefix={arXiv},
      primaryClass={cs.CL},
      url={https://arxiv.org/abs/2509.04664}, 
}

@misc{xiong2024llmsexpressuncertaintyempirical,
      title={Can LLMs Express Their Uncertainty? An Empirical Evaluation of Confidence Elicitation in LLMs}, 
      author={Miao Xiong and Zhiyuan Hu and Xinyang Lu and Yifei Li and Jie Fu and Junxian He and Bryan Hooi},
      year={2024},
      eprint={2306.13063},
      archivePrefix={arXiv},
      primaryClass={cs.CL},
      url={https://arxiv.org/abs/2306.13063}, 
}

@article{griot25,
author = {Griot, Maxime and Hemptinne, Coralie and Vanderdonckt, Jean and Yuksel, Demet},
year = {2025},
month = {01},
pages = {},
title = {Large Language Models lack essential metacognition for reliable medical reasoning},
volume = {16},
journal = {Nature Communications},
doi = {10.1038/s41467-024-55628-6}
}

@misc{sun2025largelanguagemodelsoverconfident,
      title={Large Language Models are overconfident and amplify human bias}, 
      author={Fengfei Sun and Ningke Li and Kailong Wang and Lorenz Goette},
      year={2025},
      eprint={2505.02151},
      archivePrefix={arXiv},
      primaryClass={cs.SE},
      url={https://arxiv.org/abs/2505.02151}, 
}

@article{colombatto25,
author = {Colombatto, Clara and Birch, Jonathan and Fleming, Stephen},
year = {2025},
month = {05},
pages = {},
title = {The influence of mental state attributions on trust in large language models},
volume = {3},
journal = {Communications Psychology},
doi = {10.1038/s44271-025-00262-1}
}

@misc{wang2025objectivefinetuningllmsprior,
      title={Towards Objective Fine-tuning: How LLMs' Prior Knowledge Causes Potential Poor Calibration?}, 
      author={Ziming Wang and Zeyu Shi and Haoyi Zhou and Shiqi Gao and Qingyun Sun and Jianxin Li},
      year={2025},
      eprint={2505.20903},
      archivePrefix={arXiv},
      primaryClass={cs.CL},
      url={https://arxiv.org/abs/2505.20903}, 
}

@inproceedings{kapoor-etal-2024-calibration,
    title = "Calibration-Tuning: Teaching Large Language Models to Know What They Don{'}t Know",
    author = "Kapoor, Sanyam  and
      Gruver, Nate  and
      Roberts, Manley  and
      Pal, Arka  and
      Dooley, Samuel  and
      Goldblum, Micah  and
      Wilson, Andrew",
    editor = {V{\'a}zquez, Ra{\'u}l  and
      Celikkanat, Hande  and
      Ulmer, Dennis  and
      Tiedemann, J{\"o}rg  and
      Swayamdipta, Swabha  and
      Aziz, Wilker  and
      Plank, Barbara  and
      Baan, Joris  and
      de Marneffe, Marie-Catherine},
    booktitle = "Proceedings of the 1st Workshop on Uncertainty-Aware NLP (UncertaiNLP 2024)",
    month = mar,
    year = "2024",
    address = "St Julians, Malta",
    publisher = "Association for Computational Linguistics",
    url = "https://aclanthology.org/2024.uncertainlp-1.1/",
    pages = "1--14"
}

@misc{qwen2025qwen25technicalreport,
      title={Qwen2.5 Technical Report}, 
      author={Qwen and : and An Yang and Baosong Yang and Beichen Zhang and Binyuan Hui and Bo Zheng and Bowen Yu and Chengyuan Li and Dayiheng Liu and Fei Huang and Haoran Wei and Huan Lin and Jian Yang and Jianhong Tu and Jianwei Zhang and Jianxin Yang and Jiaxi Yang and Jingren Zhou and Junyang Lin and Kai Dang and Keming Lu and Keqin Bao and Kexin Yang and Le Yu and Mei Li and Mingfeng Xue and Pei Zhang and Qin Zhu and Rui Men and Runji Lin and Tianhao Li and Tianyi Tang and Tingyu Xia and Xingzhang Ren and Xuancheng Ren and Yang Fan and Yang Su and Yichang Zhang and Yu Wan and Yuqiong Liu and Zeyu Cui and Zhenru Zhang and Zihan Qiu},
      year={2025},
      eprint={2412.15115},
      archivePrefix={arXiv},
      primaryClass={cs.CL},
      url={https://arxiv.org/abs/2412.15115}, 
}

@misc{grattafiori2024llama3herdmodels,
      title={The Llama 3 Herd of Models}, 
      author={Aaron Grattafiori and Abhimanyu Dubey and Abhinav Jauhri and Abhinav Pandey and Abhishek Kadian and Ahmad Al-Dahle and Aiesha Letman and Akhil Mathur and Alan Schelten and Alex Vaughan and Amy Yang and Angela Fan and Anirudh Goyal and Anthony Hartshorn and Aobo Yang and Archi Mitra and Archie Sravankumar and Artem Korenev and Arthur Hinsvark and Arun Rao and Aston Zhang and Aurelien Rodriguez and Austen Gregerson and Ava Spataru and Baptiste Roziere and Bethany Biron and Binh Tang and Bobbie Chern and Charlotte Caucheteux and Chaya Nayak and Chloe Bi and Chris Marra and Chris McConnell and Christian Keller and Christophe Touret and Chunyang Wu and Corinne Wong and Cristian Canton Ferrer and Cyrus Nikolaidis and Damien Allonsius and Daniel Song and Danielle Pintz and Danny Livshits and Danny Wyatt and David Esiobu and Dhruv Choudhary and Dhruv Mahajan and Diego Garcia-Olano and Diego Perino and Dieuwke Hupkes and Egor Lakomkin and Ehab AlBadawy and Elina Lobanova and Emily Dinan and Eric Michael Smith and Filip Radenovic and Francisco Guzmán and Frank Zhang and Gabriel Synnaeve and Gabrielle Lee and Georgia Lewis Anderson and Govind Thattai and Graeme Nail and Gregoire Mialon and Guan Pang and Guillem Cucurell and Hailey Nguyen and Hannah Korevaar and Hu Xu and Hugo Touvron and Iliyan Zarov and Imanol Arrieta Ibarra and Isabel Kloumann and Ishan Misra and Ivan Evtimov and Jack Zhang and Jade Copet and Jaewon Lee and Jan Geffert and Jana Vranes and Jason Park and Jay Mahadeokar and Jeet Shah and Jelmer van der Linde and Jennifer Billock and Jenny Hong and Jenya Lee and Jeremy Fu and Jianfeng Chi and Jianyu Huang and Jiawen Liu and Jie Wang and Jiecao Yu and Joanna Bitton and Joe Spisak and Jongsoo Park and Joseph Rocca and Joshua Johnstun and Joshua Saxe and Junteng Jia and Kalyan Vasuden Alwala and Karthik Prasad and Kartikeya Upasani and Kate Plawiak and Ke Li and Kenneth Heafield and Kevin Stone and Khalid El-Arini and Krithika Iyer and Kshitiz Malik and Kuenley Chiu and Kunal Bhalla and Kushal Lakhotia and Lauren Rantala-Yeary and Laurens van der Maaten and Lawrence Chen and Liang Tan and Liz Jenkins and Louis Martin and Lovish Madaan and Lubo Malo and Lukas Blecher and Lukas Landzaat and Luke de Oliveira and Madeline Muzzi and Mahesh Pasupuleti and Mannat Singh and Manohar Paluri and Marcin Kardas and Maria Tsimpoukelli and Mathew Oldham and Mathieu Rita and Maya Pavlova and Melanie Kambadur and Mike Lewis and Min Si and Mitesh Kumar Singh and Mona Hassan and Naman Goyal and Narjes Torabi and Nikolay Bashlykov and Nikolay Bogoychev and Niladri Chatterji and Ning Zhang and Olivier Duchenne and Onur Çelebi and Patrick Alrassy and Pengchuan Zhang and Pengwei Li and Petar Vasic and Peter Weng and Prajjwal Bhargava and Pratik Dubal and Praveen Krishnan and Punit Singh Koura and Puxin Xu and Qing He and Qingxiao Dong and Ragavan Srinivasan and Raj Ganapathy and Ramon Calderer and Ricardo Silveira Cabral and Robert Stojnic and Roberta Raileanu and Rohan Maheswari and Rohit Girdhar and Rohit Patel and Romain Sauvestre and Ronnie Polidoro and Roshan Sumbaly and Ross Taylor and Ruan Silva and Rui Hou and Rui Wang and Saghar Hosseini and Sahana Chennabasappa and Sanjay Singh and Sean Bell and Seohyun Sonia Kim and Sergey Edunov and Shaoliang Nie and Sharan Narang and Sharath Raparthy and Sheng Shen and Shengye Wan and Shruti Bhosale and Shun Zhang and Simon Vandenhende and Soumya Batra and Spencer Whitman and Sten Sootla and Stephane Collot and Suchin Gururangan and Sydney Borodinsky and Tamar Herman and Tara Fowler and Tarek Sheasha and Thomas Georgiou and Thomas Scialom and Tobias Speckbacher and Todor Mihaylov and Tong Xiao and Ujjwal Karn and Vedanuj Goswami and Vibhor Gupta and Vignesh Ramanathan and Viktor Kerkez and Vincent Gonguet and Virginie Do and Vish Vogeti and Vítor Albiero and Vladan Petrovic and Weiwei Chu and Wenhan Xiong and Wenyin Fu and Whitney Meers and Xavier Martinet and Xiaodong Wang and Xiaofang Wang and Xiaoqing Ellen Tan and Xide Xia and Xinfeng Xie and Xuchao Jia and Xuewei Wang and Yaelle Goldschlag and Yashesh Gaur and Yasmine Babaei and Yi Wen and Yiwen Song and Yuchen Zhang and Yue Li and Yuning Mao and Zacharie Delpierre Coudert and Zheng Yan and Zhengxing Chen and Zoe Papakipos and Aaditya Singh and Aayushi Srivastava and Abha Jain and Adam Kelsey and Adam Shajnfeld and Adithya Gangidi and Adolfo Victoria and Ahuva Goldstand and Ajay Menon and Ajay Sharma and Alex Boesenberg and Alexei Baevski and Allie Feinstein and Amanda Kallet and Amit Sangani and Amos Teo and Anam Yunus and Andrei Lupu and Andres Alvarado and Andrew Caples and Andrew Gu and Andrew Ho and Andrew Poulton and Andrew Ryan and Ankit Ramchandani and Annie Dong and Annie Franco and Anuj Goyal and Aparajita Saraf and Arkabandhu Chowdhury and Ashley Gabriel and Ashwin Bharambe and Assaf Eisenman and Azadeh Yazdan and Beau James and Ben Maurer and Benjamin Leonhardi and Bernie Huang and Beth Loyd and Beto De Paola and Bhargavi Paranjape and Bing Liu and Bo Wu and Boyu Ni and Braden Hancock and Bram Wasti and Brandon Spence and Brani Stojkovic and Brian Gamido and Britt Montalvo and Carl Parker and Carly Burton and Catalina Mejia and Ce Liu and Changhan Wang and Changkyu Kim and Chao Zhou and Chester Hu and Ching-Hsiang Chu and Chris Cai and Chris Tindal and Christoph Feichtenhofer and Cynthia Gao and Damon Civin and Dana Beaty and Daniel Kreymer and Daniel Li and David Adkins and David Xu and Davide Testuggine and Delia David and Devi Parikh and Diana Liskovich and Didem Foss and Dingkang Wang and Duc Le and Dustin Holland and Edward Dowling and Eissa Jamil and Elaine Montgomery and Eleonora Presani and Emily Hahn and Emily Wood and Eric-Tuan Le and Erik Brinkman and Esteban Arcaute and Evan Dunbar and Evan Smothers and Fei Sun and Felix Kreuk and Feng Tian and Filippos Kokkinos and Firat Ozgenel and Francesco Caggioni and Frank Kanayet and Frank Seide and Gabriela Medina Florez and Gabriella Schwarz and Gada Badeer and Georgia Swee and Gil Halpern and Grant Herman and Grigory Sizov and Guangyi and Zhang and Guna Lakshminarayanan and Hakan Inan and Hamid Shojanazeri and Han Zou and Hannah Wang and Hanwen Zha and Haroun Habeeb and Harrison Rudolph and Helen Suk and Henry Aspegren and Hunter Goldman and Hongyuan Zhan and Ibrahim Damlaj and Igor Molybog and Igor Tufanov and Ilias Leontiadis and Irina-Elena Veliche and Itai Gat and Jake Weissman and James Geboski and James Kohli and Janice Lam and Japhet Asher and Jean-Baptiste Gaya and Jeff Marcus and Jeff Tang and Jennifer Chan and Jenny Zhen and Jeremy Reizenstein and Jeremy Teboul and Jessica Zhong and Jian Jin and Jingyi Yang and Joe Cummings and Jon Carvill and Jon Shepard and Jonathan McPhie and Jonathan Torres and Josh Ginsburg and Junjie Wang and Kai Wu and Kam Hou U and Karan Saxena and Kartikay Khandelwal and Katayoun Zand and Kathy Matosich and Kaushik Veeraraghavan and Kelly Michelena and Keqian Li and Kiran Jagadeesh and Kun Huang and Kunal Chawla and Kyle Huang and Lailin Chen and Lakshya Garg and Lavender A and Leandro Silva and Lee Bell and Lei Zhang and Liangpeng Guo and Licheng Yu and Liron Moshkovich and Luca Wehrstedt and Madian Khabsa and Manav Avalani and Manish Bhatt and Martynas Mankus and Matan Hasson and Matthew Lennie and Matthias Reso and Maxim Groshev and Maxim Naumov and Maya Lathi and Meghan Keneally and Miao Liu and Michael L. Seltzer and Michal Valko and Michelle Restrepo and Mihir Patel and Mik Vyatskov and Mikayel Samvelyan and Mike Clark and Mike Macey and Mike Wang and Miquel Jubert Hermoso and Mo Metanat and Mohammad Rastegari and Munish Bansal and Nandhini Santhanam and Natascha Parks and Natasha White and Navyata Bawa and Nayan Singhal and Nick Egebo and Nicolas Usunier and Nikhil Mehta and Nikolay Pavlovich Laptev and Ning Dong and Norman Cheng and Oleg Chernoguz and Olivia Hart and Omkar Salpekar and Ozlem Kalinli and Parkin Kent and Parth Parekh and Paul Saab and Pavan Balaji and Pedro Rittner and Philip Bontrager and Pierre Roux and Piotr Dollar and Polina Zvyagina and Prashant Ratanchandani and Pritish Yuvraj and Qian Liang and Rachad Alao and Rachel Rodriguez and Rafi Ayub and Raghotham Murthy and Raghu Nayani and Rahul Mitra and Rangaprabhu Parthasarathy and Raymond Li and Rebekkah Hogan and Robin Battey and Rocky Wang and Russ Howes and Ruty Rinott and Sachin Mehta and Sachin Siby and Sai Jayesh Bondu and Samyak Datta and Sara Chugh and Sara Hunt and Sargun Dhillon and Sasha Sidorov and Satadru Pan and Saurabh Mahajan and Saurabh Verma and Seiji Yamamoto and Sharadh Ramaswamy and Shaun Lindsay and Shaun Lindsay and Sheng Feng and Shenghao Lin and Shengxin Cindy Zha and Shishir Patil and Shiva Shankar and Shuqiang Zhang and Shuqiang Zhang and Sinong Wang and Sneha Agarwal and Soji Sajuyigbe and Soumith Chintala and Stephanie Max and Stephen Chen and Steve Kehoe and Steve Satterfield and Sudarshan Govindaprasad and Sumit Gupta and Summer Deng and Sungmin Cho and Sunny Virk and Suraj Subramanian and Sy Choudhury and Sydney Goldman and Tal Remez and Tamar Glaser and Tamara Best and Thilo Koehler and Thomas Robinson and Tianhe Li and Tianjun Zhang and Tim Matthews and Timothy Chou and Tzook Shaked and Varun Vontimitta and Victoria Ajayi and Victoria Montanez and Vijai Mohan and Vinay Satish Kumar and Vishal Mangla and Vlad Ionescu and Vlad Poenaru and Vlad Tiberiu Mihailescu and Vladimir Ivanov and Wei Li and Wenchen Wang and Wenwen Jiang and Wes Bouaziz and Will Constable and Xiaocheng Tang and Xiaojian Wu and Xiaolan Wang and Xilun Wu and Xinbo Gao and Yaniv Kleinman and Yanjun Chen and Ye Hu and Ye Jia and Ye Qi and Yenda Li and Yilin Zhang and Ying Zhang and Yossi Adi and Youngjin Nam and Yu and Wang and Yu Zhao and Yuchen Hao and Yundi Qian and Yunlu Li and Yuzi He and Zach Rait and Zachary DeVito and Zef Rosnbrick and Zhaoduo Wen and Zhenyu Yang and Zhiwei Zhao and Zhiyu Ma},
      year={2024},
      eprint={2407.21783},
      archivePrefix={arXiv},
      primaryClass={cs.AI},
      url={https://arxiv.org/abs/2407.21783}, 
}

@misc{abdin2024phi4technicalreport,
      title={Phi-4 Technical Report}, 
      author={Marah Abdin and Jyoti Aneja and Harkirat Behl and Sébastien Bubeck and Ronen Eldan and Suriya Gunasekar and Michael Harrison and Russell J. Hewett and Mojan Javaheripi and Piero Kauffmann and James R. Lee and Yin Tat Lee and Yuanzhi Li and Weishung Liu and Caio C. T. Mendes and Anh Nguyen and Eric Price and Gustavo de Rosa and Olli Saarikivi and Adil Salim and Shital Shah and Xin Wang and Rachel Ward and Yue Wu and Dingli Yu and Cyril Zhang and Yi Zhang},
      year={2024},
      eprint={2412.08905},
      archivePrefix={arXiv},
      primaryClass={cs.CL},
      url={https://arxiv.org/abs/2412.08905}, 
}

@misc{jiang2023mistral7b,
      title={Mistral 7B}, 
      author={Albert Q. Jiang and Alexandre Sablayrolles and Arthur Mensch and Chris Bamford and Devendra Singh Chaplot and Diego de las Casas and Florian Bressand and Gianna Lengyel and Guillaume Lample and Lucile Saulnier and Lélio Renard Lavaud and Marie-Anne Lachaux and Pierre Stock and Teven Le Scao and Thibaut Lavril and Thomas Wang and Timothée Lacroix and William El Sayed},
      year={2023},
      eprint={2310.06825},
      archivePrefix={arXiv},
      primaryClass={cs.CL},
      url={https://arxiv.org/abs/2310.06825}, 
}

@misc{kuhn2023semanticuncertaintylinguisticinvariances,
      title={Semantic Uncertainty: Linguistic Invariances for Uncertainty Estimation in Natural Language Generation}, 
      author={Lorenz Kuhn and Yarin Gal and Sebastian Farquhar},
      year={2023},
      eprint={2302.09664},
      archivePrefix={arXiv},
      primaryClass={cs.CL},
      url={https://arxiv.org/abs/2302.09664}, 
}

@article{caziot2021perceptual,
  title={Perceptual confidence judgments reflect self-consistency},
  author={Caziot, Baptiste and Mamassian, Pascal},
  journal={Journal of Vision},
  volume={21},
  number={12},
  pages={8--8},
  year={2021},
  publisher={The Association for Research in Vision and Ophthalmology},
  doi={10.1167/jov.21.12.8},
  url={https://doi.org/10.1167/jov.21.12.8}
}

@misc{kumaran2026reportedconfidencellmstracks,
      title={Reported Confidence in LLMs Tracks Commitment More Than Correctness}, 
      author={Dharshan Kumaran},
      year={2026},
      eprint={2606.29490},
      archivePrefix={arXiv},
      primaryClass={cs.LG},
      url={https://arxiv.org/abs/2606.29490}, 
}

@misc{zhang2025qwen3embeddingadvancingtext,
      title={Qwen3 Embedding: Advancing Text Embedding and Reranking Through Foundation Models}, 
      author={Yanzhao Zhang and Mingxin Li and Dingkun Long and Xin Zhang and Huan Lin and Baosong Yang and Pengjun Xie and An Yang and Dayiheng Liu and Junyang Lin and Fei Huang and Jingren Zhou},
      year={2025},
      eprint={2506.05176},
      archivePrefix={arXiv},
      primaryClass={cs.CL},
      url={https://arxiv.org/abs/2506.05176}, 
}

@misc{manakul2023selfcheckgptzeroresourceblackboxhallucination,
      title={SelfCheckGPT: Zero-Resource Black-Box Hallucination Detection for Generative Large Language Models}, 
      author={Potsawee Manakul and Adian Liusie and Mark J. F. Gales},
      year={2023},
      eprint={2303.08896},
      archivePrefix={arXiv},
      primaryClass={cs.CL},
      url={https://arxiv.org/abs/2303.08896}, 
}

@misc{yang2025qwen3technicalreport,
      title={Qwen3 Technical Report}, 
      author={An Yang and Anfeng Li and Baosong Yang and Beichen Zhang and Binyuan Hui and Bo Zheng and Bowen Yu and Chang Gao and Chengen Huang and Chenxu Lv and Chujie Zheng and Dayiheng Liu and Fan Zhou and Fei Huang and Feng Hu and Hao Ge and Haoran Wei and Huan Lin and Jialong Tang and Jian Yang and Jianhong Tu and Jianwei Zhang and Jianxin Yang and Jiaxi Yang and Jing Zhou and Jingren Zhou and Junyang Lin and Kai Dang and Keqin Bao and Kexin Yang and Le Yu and Lianghao Deng and Mei Li and Mingfeng Xue and Mingze Li and Pei Zhang and Peng Wang and Qin Zhu and Rui Men and Ruize Gao and Shixuan Liu and Shuang Luo and Tianhao Li and Tianyi Tang and Wenbiao Yin and Xingzhang Ren and Xinyu Wang and Xinyu Zhang and Xuancheng Ren and Yang Fan and Yang Su and Yichang Zhang and Yinger Zhang and Yu Wan and Yuqiong Liu and Zekun Wang and Zeyu Cui and Zhenru Zhang and Zhipeng Zhou and Zihan Qiu},
      year={2025},
      eprint={2505.09388},
      archivePrefix={arXiv},
      primaryClass={cs.CL},
      url={https://arxiv.org/abs/2505.09388}, 
}

@misc{touvron2023llama2openfoundation,
      title={Llama 2: Open Foundation and Fine-Tuned Chat Models}, 
      author={Hugo Touvron and Louis Martin and Kevin Stone and Peter Albert and Amjad Almahairi and Yasmine Babaei and Nikolay Bashlykov and Soumya Batra and Prajjwal Bhargava and Shruti Bhosale and Dan Bikel and Lukas Blecher and Cristian Canton Ferrer and Moya Chen and Guillem Cucurull and David Esiobu and Jude Fernandes and Jeremy Fu and Wenyin Fu and Brian Fuller and Cynthia Gao and Vedanuj Goswami and Naman Goyal and Anthony Hartshorn and Saghar Hosseini and Rui Hou and Hakan Inan and Marcin Kardas and Viktor Kerkez and Madian Khabsa and Isabel Kloumann and Artem Korenev and Punit Singh Koura and Marie-Anne Lachaux and Thibaut Lavril and Jenya Lee and Diana Liskovich and Yinghai Lu and Yuning Mao and Xavier Martinet and Todor Mihaylov and Pushkar Mishra and Igor Molybog and Yixin Nie and Andrew Poulton and Jeremy Reizenstein and Rashi Rungta and Kalyan Saladi and Alan Schelten and Ruan Silva and Eric Michael Smith and Ranjan Subramanian and Xiaoqing Ellen Tan and Binh Tang and Ross Taylor and Adina Williams and Jian Xiang Kuan and Puxin Xu and Zheng Yan and Iliyan Zarov and Yuchen Zhang and Angela Fan and Melanie Kambadur and Sharan Narang and Aurelien Rodriguez and Robert Stojnic and Sergey Edunov and Thomas Scialom},
      year={2023},
      eprint={2307.09288},
      archivePrefix={arXiv},
      primaryClass={cs.CL},
      url={https://arxiv.org/abs/2307.09288}, 
}

@misc{liu2026ministral3,
      title={Ministral 3}, 
      author={Alexander H. Liu and Kartik Khandelwal and Sandeep Subramanian and Victor Jouault and Abhinav Rastogi and Adrien Sadé and Alan Jeffares and Albert Jiang and Alexandre Cahill and Alexandre Gavaudan and Alexandre Sablayrolles and Amélie Héliou and Amos You and Andy Ehrenberg and Andy Lo and Anton Eliseev and Antonia Calvi and Avinash Sooriyarachchi and Baptiste Bout and Baptiste Rozière and Baudouin De Monicault and Clémence Lanfranchi and Corentin Barreau and Cyprien Courtot and Daniele Grattarola and Darius Dabert and Diego de las Casas and Elliot Chane-Sane and Faruk Ahmed and Gabrielle Berrada and Gaëtan Ecrepont and Gauthier Guinet and Georgii Novikov and Guillaume Kunsch and Guillaume Lample and Guillaume Martin and Gunshi Gupta and Jan Ludziejewski and Jason Rute and Joachim Studnia and Jonas Amar and Joséphine Delas and Josselin Somerville Roberts and Karmesh Yadav and Khyathi Chandu and Kush Jain and Laurence Aitchison and Laurent Fainsin and Léonard Blier and Lingxiao Zhao and Louis Martin and Lucile Saulnier and Luyu Gao and Maarten Buyl and Margaret Jennings and Marie Pellat and Mark Prins and Mathieu Poirée and Mathilde Guillaumin and Matthieu Dinot and Matthieu Futeral and Maxime Darrin and Maximilian Augustin and Mia Chiquier and Michel Schimpf and Nathan Grinsztajn and Neha Gupta and Nikhil Raghuraman and Olivier Bousquet and Olivier Duchenne and Patricia Wang and Patrick von Platen and Paul Jacob and Paul Wambergue and Paula Kurylowicz and Pavankumar Reddy Muddireddy and Philomène Chagniot and Pierre Stock and Pravesh Agrawal and Quentin Torroba and Romain Sauvestre and Roman Soletskyi and Rupert Menneer and Sagar Vaze and Samuel Barry and Sanchit Gandhi and Siddhant Waghjale and Siddharth Gandhi and Soham Ghosh and Srijan Mishra and Sumukh Aithal and Szymon Antoniak and Teven Le Scao and Théo Cachet and Theo Simon Sorg and Thibaut Lavril and Thiziri Nait Saada and Thomas Chabal and Thomas Foubert and Thomas Robert and Thomas Wang and Tim Lawson and Tom Bewley and Tom Bewley and Tom Edwards and Umar Jamil and Umberto Tomasini and Valeriia Nemychnikova and Van Phung and Vincent Maladière and Virgile Richard and Wassim Bouaziz and Wen-Ding Li and William Marshall and Xinghui Li and Xinyu Yang and Yassine El Ouahidi and Yihan Wang and Yunhao Tang and Zaccharie Ramzi},
      year={2026},
      eprint={2601.08584},
      archivePrefix={arXiv},
      primaryClass={cs.CL},
      url={https://arxiv.org/abs/2601.08584}, 
}

@misc{unsloth,
  author = {UnslothAI and Han-Chen, Daniel and Han-Chen, Michael},
  title = {Unsloth},
  year = {2025},
  publisher = {GitHub},
  howpublished = {\url{https://github.com/unslothai/unsloth}}
}

@misc{weidinger2021ethicalsocialrisksharm,
      title={Ethical and social risks of harm from Language Models}, 
      author={Laura Weidinger and John Mellor and Maribeth Rauh and Conor Griffin and Jonathan Uesato and Po-Sen Huang and Myra Cheng and Mia Glaese and Borja Balle and Atoosa Kasirzadeh and Zac Kenton and Sasha Brown and Will Hawkins and Tom Stepleton and Courtney Biles and Abeba Birhane and Julia Haas and Laura Rimell and Lisa Anne Hendricks and William Isaac and Sean Legassick and Geoffrey Irving and Iason Gabriel},
      year={2021},
      eprint={2112.04359},
      archivePrefix={arXiv},
      primaryClass={cs.CL},
      url={https://arxiv.org/abs/2112.04359}, 
}

@article{steyvers2025large,
  title   = {What large language models know and what people think they know},
  author  = {Steyvers, Mark and Tejeda, Heliodoro and Kumar, Aakriti and Belem, Catarina and Karny, Sheer and Hu, Xinyue and Mayer, Lukas W. and Smyth, Padhraic},
  journal = {Nature Machine Intelligence},
  volume  = {7},
  number  = {2},
  pages   = {221--231},
  year    = {2025},
  doi     = {10.1038/s42256-024-00976-7}
}
\bibliographystyle{iclr2026_conference}

\appendix
\clearpage
\section*{Appendix}
\addcontentsline{toc}{section}{Appendix overview}

\noindent\textbf{Data and experimental setup}
\begin{itemize}
    \item[\ref{app:math_dataset}] Custom \textsc{Math} dataset: question generation, difficulty levels, categories, and distractor design.
    \item[\ref{app:question_formatting}] Question formatting: answer and confidence prompts, how option probabilities are read out, and how verbalized confidence is parsed.
    \item[\ref{app:dataset_splitting}] Dataset splitting: category-level train/test split procedure, resulting categories, and per-model accuracy on each split.
    \item[\ref{app:hyperparameters}] Hyperparameters: training settings shared across all models and datasets.
\end{itemize}

\noindent\textbf{Additional results}
\begin{itemize}
    \item[\ref{sec:development}] Emergence during training: how output consistency tracking and accuracy tracking develop across training checkpoints.
    \item[\ref{app:middle_probe}] Middle probe: replication of all main analyses with the probe at the middle of the network, ruling out a logit-leak explanation.
\end{itemize}

\noindent\textbf{Supplementary figures}
\begin{itemize}
    \item[\ref{app:training_curves}] Training curves: train and test loss curves during training for both end and middle probes.
    \item[\ref{app:individual_plots}] Individual model plots: per-model confidence performance before and after training, and cross-dataset correlations and $\Delta r$ for both probes.
\end{itemize}

\clearpage
\section{Custom MATH dataset}
\label{app:math_dataset}

The \textsc{Math} datasets consist of procedurally generated arithmetic questions of the form ``What is $a \circ b$?'', with $\circ \in \{+, -, \times\}$. Answering them does not require any world knowledge, and difficulty can be controlled precisely through the magnitude of the operands. This gives a set of question categories that are thematically more homogeneous, in contrast to \textsc{MMLU-Pro} and \textsc{MedMCQA} while remaining challenging and distinct questions.

\paragraph{Operands and difficulty levels.}
We generate $120\,000$ questions across three difficulty levels, defined by the number of digits of the smaller operand $b$. The larger operand $a$ is written first. Pairs $(a, b)$ are sampled uniformly without replacement from the ranges below, so no question appears twice within a level:
\begin{itemize}
    \item \textbf{easy} (30\% of questions): $b \in [1, 9]$, $a \in [10, 99\,999]$;
    \item \textbf{medium} (40\%): $b \in [10, 99]$, $a \in [100, 99\,999]$;
    \item \textbf{hard} (30\%): $b \in [100, 999]$, $a \in [1\,000, 99\,999]$.
\end{itemize}
Within each level, the three operators are assigned in rotation, so each operator covers exactly one third of the level. Because $a > b$ in every range, results of subtractions are always strictly positive. Each combination of level and operator defines a category (e.g., medium$*$ for medium multiplications), which yields the 9 categories listed in Table~\ref{tab:math_categories}. Following the procedure of Appendix~\ref{app:dataset_splitting}, the category easy$*$ is held out as the test split and the 8 remaining categories form the training split. This gives $108\,000$ training and $12\,000$ test questions.

\begin{table}[h]
\centering
\small
\begin{tabular}{llrrl}
\toprule
Level & Operators & $b$ range & $a$ range & \#questions per operator \\
\midrule
easy   & $+, -, \times$ & $[1, 9]$       & $[10, 99\,999]$    & $12\,000$ \\
medium & $+, -, \times$ & $[10, 99]$     & $[100, 99\,999]$   & $16\,000$ \\
hard   & $+, -, \times$ & $[100, 999]$   & $[1\,000, 99\,999]$ & $12\,000$ \\
\bottomrule
\end{tabular}
\caption{Categories of the \textsc{Math} datasets. Each level is split evenly between the three operators, yielding 9 categories (e.g., easy$+$, easy$-$, easy$*$).}
\label{tab:math_categories}
\end{table}

\paragraph{Distractor generation.}
Distractors are designed so that questions cannot be solved by shallow heuristics such as checking the magnitude of the result or its last digit. Each distractor is obtained by altering some digits of the correct result $r$, whose decimal representation has $L$ digits:
\begin{enumerate}
    \item \textbf{Candidate positions.} If $L > 2$, the first and last digits of $r$ are kept fixed and only the inner positions $\{1, \dots, L-2\}$ may be altered. If $L \le 2$, all positions may be altered.
    \item \textbf{Number of altered digits.} We draw $n = \min\left(1 + k,\ 4,\ |\text{candidates}|\right)$ with $k \sim \mathrm{Poisson}\left(\tfrac{L-1}{2}\right)$. Longer results therefore tend to have more altered digits, up to at most 4.
    \item \textbf{Shared positions.} A set of $n$ distinct positions is sampled uniformly among the candidates. The same positions are used for all distractors of a question.
    \item \textbf{Distractor values.} For each distractor, the digits at these positions are replaced by uniformly random digits. We reject combinations that equal the correct result's digits at these positions and combinations already used by another distractor. This guarantees that all options are distinct.
\end{enumerate}
As a result, all options of a question share the same leading digit, the same last digit and, in general, the same number of digits. They differ only at the altered positions. For example, the medium$*$ question ``What is $48213 * 87$?'' ($r = 4\,194\,531$) could receive the altered positions $\{2, 4\}$ and the options $\{4\,194\,531,\ 4\,134\,231,\ 4\,194\,831,\ 4\,164\,531\}$. Note that a single distractor may keep some of the correct digits, since only the full combination at the altered positions must differ.

The correct answer is inserted among the $M_\mathcal{D} - 1$ distractors, and the options are shuffled uniformly, so the position of the correct answer is uniformly distributed.

\paragraph{4- and 10-option variants.}
\textsc{Math}$_4$ and \textsc{Math}$_{10}$ are generated with the same random seed. Operand pairs are sampled before any distractor is drawn, so the two variants contain exactly the same questions and category split. They differ only in their number of options (3 and 9 distractors, respectively), and thus in their distractors.

\clearpage
\section{Question formatting}
\label{app:question_formatting}

All models are evaluated zero-shot, with no few-shot examples, no system prompt and no chat template. Instruction-tuned models therefore receive the same raw text as base models. In every prompt, the question $q$ is introduced by \texttt{"Question :"}, so that $\squest{} = \texttt{"Question :"} \cat q$. It is followed by the options, one per line. Options are labelled with capital letters in their dataset order: (A) to (D) for 4-option variants and (A) to (J) for 10-option variants. The options block is $\sopt[] = \sopt[0] \cdots \sopt[M_\mathcal{D}-1]$, where $\sopt[j] = \texttt{"\textbackslash n("} \cat L_j \cat \texttt{") "} \cat o_j$ contains the letter $L_j$ and the text $o_j$ of option $j$. The prompt then ends with a cue line, which differs between answering and confidence elicitation. For readability, the cue line is left implicit in the notation of the main text.

\paragraph{Answer prompt.}
To evaluate true accuracy and output consistency, the options are followed by the cue \texttt{"\textbackslash nAnswer : Option ("}, so that the next token the model produces is the letter of its chosen option. For example, a \textsc{Math}$_4$ item is formatted as:
\begin{verbatim}
Question :What is 48213 * 87?
(A) 4134231
(B) 4194531
(C) 4194831
(D) 4164531
Answer : Option (
\end{verbatim}

\paragraph{Reading out the answer distribution.}
We do not sample answers. We rather run a single forward pass on the prompt and read the next-token distribution $p_\theta(\cdot \mid \squest{}\sopt[])$ at the last position. Prompts are batched with left padding, so that the last position always corresponds to the final prompt token. Tokenizers may encode an option letter in several ways depending on the surrounding whitespace and punctuation. As such, the probability of option $j$ is the sum of the probabilities of all tokens in the vocabulary that could correspond to letter $L_j$:
\begin{equation}
p_\theta(\sopt[j] \mid \squest{}\sopt[]) = \sum_{t \in V(L_j)} p_\theta(t \mid \squest{}\sopt[]),
\end{equation}
where $V(L_j)$ contains the tokens among \texttt{"$L_j$"}, \texttt{"$L_j$ "}, \texttt{"$L_j$)"}, \texttt{"$L_j$) "}, \texttt{" $L_j$"}, \texttt{" $L_j$ "}, \texttt{" $L_j$)"} and \texttt{" $L_j$) "} that exist in the model's vocabulary and could represent the same $L_j$ answer. The accuracy of the model on an item (Section~\ref{sec:evaluation_procedures}) is the probability assigned in this way to the letter of the correct option. It is thus a soft measure of correctness: the expected accuracy of an answer sampled from the model, rather than a binary score.

\paragraph{Confidence prompt.}
To elicit verbalized confidence, the question is preceded by a pre-prompt $\spre{}$ that asks the model to estimate its probability of answering correctly and specifies the expected answer format:
\begin{quote}
\texttt{You are going to be asked to guess your likelihood to be right on a given question. Give your answer with the format 'Accuracy prediction: X\%' but replace X with your percentage chance to be right on this question.\textbackslash n}
\end{quote}
The options are then followed by the cue \texttt{"\textbackslash nAccuracy prediction:"} instead of the answer cue. The model is never asked to answer the question in this prompt. The same \textsc{Math}$_4$ item thus becomes:
\begin{verbatim}
You are going to be asked to guess your likelihood to be right on a
given question. Give your answer with the format 'Accuracy prediction:
X%' but replace X with your percentage chance to be right on this
question.
Question :What is 48213 * 87?
(A) 4134231
(B) 4194531
(C) 4194831
(D) 4164531
Accuracy prediction:
\end{verbatim}
(Line breaks inside the instruction are added here for display only.) We generate exactly 5 tokens with greedy decoding. The completion is then searched for numbers immediately followed by a percent sign (regular expression \verb|(\d+(?:\.\d+)?)%|). If exactly one such number is found, it is taken as the model's confidence, rescaled from percent to $[0, 1]$. Otherwise, the response is considered invalid and has no confidence estimate. Verbalized confidence is computed on the test split of each dataset, on the same items as those used to evaluate the trained models.


\clearpage
\section{Dataset splitting}
\label{app:dataset_splitting}

Each dataset variant is split into a training set and a test set by assigning entire categories to one side or the other, never individual questions. We pool all items of the original dataset test split before partitioning. The split should satisfy two goals: (i) the training set should contain about 90\% of the items, and (ii) every model should have a similar accuracy on both splits. The second goal ensures that differences in metacognitive performance between train and test cannot be explained by a difference in task difficulty. The splits should form a partition of the categories and not mix items from similar categories across splits which could create a form of implicit form of contamination from training to test set (except for $\textsc{MATH}$ which has implicitly similar categories). In addition, these constraint have to be respected for all models at once at best for results across models to be comparable.

This appendix will describe how this procedure is done through the optimization of an objective function described below.

\paragraph{Objective.}
Let $\mathcal{C}$ be the set of categories of a dataset, $N_c$ the number of items in category $c$, and $N = \sum_{c \in \mathcal{C}} N_c$ the total number of items. A split is defined by the set of test categories $\mathcal{T} \subset \mathcal{C}$. The remaining categories $\mathcal{C} \setminus \mathcal{T}$ form the training set. The fraction of training items is
\begin{equation}
f_{\text{train}}(\mathcal{T}) = 1 - \frac{1}{N}\sum_{c \in \mathcal{T}} N_c .
\end{equation}
For a model $\mathcal{M}_\theta$ and a set of categories $\mathcal{S}$, we write $Acc(\mathcal{M}_\theta, \mathcal{S})$ for the average accuracy (Section~\ref{sec:evaluation_procedures}) over all items in these categories, so that categories contribute in proportion to their size. We select the split that minimizes
\begin{equation}
\mathcal{L}(\mathcal{T}) = \left| f_{\text{train}}(\mathcal{T}) - 0.9 \right|
+ \lambda \, \frac{1}{|\mathcal{M}|} \sum_{\mathcal{M}_\theta \in \mathcal{M}}
\left| Acc(\mathcal{M}_\theta, \mathcal{C} \setminus \mathcal{T}) - Acc(\mathcal{M}_\theta, \mathcal{T}) \right| ,
\end{equation}
where $\mathcal{M}$ is the set of 10 models studied in this paper and $\lambda = 1$ (arbitrary choice).

\paragraph{Search procedure.}
Since the number of possible partitions grows exponentially with the number of categories, we minimize $\mathcal{L}$ by random search over $10\,000$ candidate splits (random seed 42) and keep the split with the lowest loss. To concentrate the search around the target 90/10 ratio, each candidate is generated as follows:
\begin{enumerate}
    \item Draw a target test fraction $\tau \sim \mathcal{U}(0.05, 0.15)$.
    \item Shuffle the categories uniformly at random.
    \item Add categories to $\mathcal{T}$ in the shuffled order until their cumulative share of items reaches $\tau$.
\end{enumerate}
Candidates in which either split is empty are discarded.

This procedure makes it possible to rather efficiently test various splits which could satisfy the aforementioned objective.

\paragraph{Resulting splits.}
Table~\ref{tab:category_splits} lists the selected categories, and Table~\ref{tab:model_performances} reports the resulting number of items and the accuracy of each model on both splits. For \textsc{Math}$_4$ and \textsc{Math}$_{10}$, the search was run independently on each variant and returned the same split: the category easy$*$ is held out, which contains exactly 10\% of the items. For \textsc{MMLU-Pro}, the two searches returned different splits: \{computer science, other\} for \textsc{MMLU-Pro}$_4$ and \{computer science, history, philosophy\} for \textsc{MMLU-Pro}$_{10}$. To make the two variants differ only in their number of options, we applied the \textsc{MMLU-Pro}$_{10}$ split to \textsc{MMLU-Pro}$_4$ as well, which increases the loss on \textsc{MMLU-Pro}$_4$ only slightly, from $0.0361$ to $0.0368$ and moved the difference of accuracy between train and test by around 1\%.

After optimization, the training fraction is $89.5\%$ and $89.6\%$ for \textsc{MMLU-Pro}$_4$ and \textsc{MMLU-Pro}$_{10}$, and $90.0\%$ for the other variants. The per-model absolute train--test accuracy gap, averaged over models (second term of $\mathcal{L}$), is $3.2$, $2.9$, $0.7$, $4.4$ and $3.3$ points on \textsc{MMLU-Pro}$_4$, \textsc{MMLU-Pro}$_{10}$, \textsc{MedMCQA}$_4$, \textsc{Math}$_4$ and \textsc{Math}$_{10}$, respectively. The largest individual gaps occur for the two Qwen3.5-9B models on the \textsc{Math} datasets (7 to 11 points). With only nine categories, \textsc{Math} admits few splits close to the 90/10 ratio (because of the repartition of question in each difficulty), which limits how well accuracy can be balanced for every model.

\begin{table}[t]
\centering
\footnotesize
\setlength{\tabcolsep}{4pt}
\renewcommand{\arraystretch}{1.15}
\begin{tabularx}{\columnwidth}{@{}llX@{}}
\toprule
Dataset & Split & Categories \\
\midrule
\multirow{2}{*}{\textsc{MMLU-Pro}}
 & train (11) & biology, business, chemistry, economics, engineering,
                health, law, math, other, physics, psychology \\
 & test\ \ (3) & computer science, history, philosophy \\
\midrule
\multirow{2}{*}{\textsc{MedMCQA}}
 & train (18) & anaesthesia, anatomy, biochemistry, dental, ENT, forensic medicine,
                gynaecology \& obstetrics, medicine, microbiology, orthopaedics,
                pathology, pediatrics, pharmacology, psychiatry, radiology,
                social \& preventive medicine, surgery, unknown \\
 & test\ \ (3) & ophthalmology, physiology, skin \\
\midrule
\multirow{2}{*}{\textsc{Math}}
 & train (8) & easy$-$, easy$+$, medium$-$, medium$*$, medium$+$,
               hard$-$, hard$*$, hard$+$ \\
 & test\ \ (1) & easy$*$ \\
\bottomrule
\end{tabularx}
\caption{Category-level train/test split of each dataset, selected by the procedure of Appendix~\ref{app:dataset_splitting}. The 4- and 10-option variants of \textsc{MMLU-Pro} and \textsc{Math} share the same split. Counts in parentheses are the number of categories per split.}
\label{tab:category_splits}
\end{table}

\begin{table*}[t]
\centering
\small
\setlength{\tabcolsep}{4.5pt}
\renewcommand{\arraystretch}{1.05}
\begin{tabular}{l *{10}{S[table-format=2.1]}}
\toprule
\multirow{2}{*}{\textbf{Model}}
 & \multicolumn{2}{c}{\textsc{MMLU-Pro}$_4$} & \multicolumn{2}{c}{\textsc{MMLU-Pro}$_{10}$}
 & \multicolumn{2}{c}{\textsc{MedMCQA}$_4$}  & \multicolumn{2}{c}{\textsc{Math}$_4$}
 & \multicolumn{2}{c}{\textsc{Math}$_{10}$} \\
\cmidrule(lr){2-3}\cmidrule(lr){4-5}\cmidrule(lr){6-7}\cmidrule(lr){8-9}\cmidrule(lr){10-11}
 & {train} & {test} & {train} & {test} & {train} & {test} & {train} & {test} & {train} & {test} \\
\midrule
\#items ($N_\mathcal{D}$)
 & \multicolumn{2}{c}{10\,685\,/\,1\,256} & \multicolumn{2}{c}{10\,337\,/\,1\,202}
 & \multicolumn{2}{c}{108\,642\,/\,12\,123} & \multicolumn{2}{c}{108\,000\,/\,12\,000}
 & \multicolumn{2}{c}{108\,000\,/\,12\,000} \\
\midrule
Llama-2-7b-chat            & 35.5 & 33.1 & 18.3 & 17.3 & 37.8 & 36.0 & 31.8 & 32.2 & 16.3 & 19.2 \\
Llama-3.2-3B-Instruct      & 42.1 & 41.8 & 25.6 & 25.5 & 66.9 & 65.6 & 34.1 & 28.6 & 13.7 & 12.8 \\
Mistral-7B-Instruct-v0.3   & 46.1 & 48.1 & 28.3 & 29.0 & 52.1 & 50.6 & 31.2 & 26.9 & 12.8 & 13.5 \\
Ministral-3-3B-Instruct    & 46.3 & 47.2 & 31.1 & 31.9 & 49.7 & 49.8 & 37.1 & 32.5 & 15.8 & 16.1 \\
Ministral-3-8B-Instruct    & 55.1 & 58.2 & 37.6 & 41.7 & 58.8 & 58.9 & 81.1 & 78.5 & 46.8 & 38.3 \\
phi-4                      & 61.1 & 67.9 & 46.6 & 54.1 & 68.7 & 68.6 & 52.7 & 52.9 & 25.5 & 28.4 \\
Qwen2.5-7B                 & 44.5 & 47.7 & 26.6 & 29.3 & 46.1 & 45.3 & 31.9 & 34.8 & 14.6 & 16.4 \\
Qwen2.5-7B-Instruct        & 55.4 & 59.8 & 37.8 & 41.3 & 59.2 & 58.2 & 52.3 & 50.1 & 29.7 & 29.8 \\
Qwen3.5-9B-Base            & 50.4 & 54.6 & 33.6 & 37.3 & 54.1 & 54.0 & 54.1 & 43.9 & 30.0 & 22.6 \\
Qwen3.5-9B                 & 52.2 & 56.5 & 33.9 & 38.6 & 57.0 & 56.6 & 55.9 & 44.9 & 29.1 & 21.8 \\
\midrule
\textbf{Mean}              & \bfseries 48.9 & \bfseries 51.5 & \bfseries 31.9 & \bfseries 34.6
                           & \bfseries 55.0 & \bfseries 54.3 & \bfseries 46.2 & \bfseries 42.5
                           & \bfseries 23.4 & \bfseries 21.9 \\
$|\Delta|$ (train, test)
 & \multicolumn{2}{c}{$2.6$} & \multicolumn{2}{c}{$2.7$} & \multicolumn{2}{c}{$0.7$}
 & \multicolumn{2}{c}{$3.7$} & \multicolumn{2}{c}{$1.5$} \\
\bottomrule
\end{tabular}
\caption{Accuracy (\%) of the evaluated models on the train and test splits of each dataset variant. \textbf{Mean} averages over the 10 models, and $|\Delta|$ is the absolute difference between the train and test means. This is a looser measure than the per-model gap minimized by the splitting objective (Appendix~\ref{app:dataset_splitting}). \#items is given as train\,/\,test. \textsc{MMLU-Pro}$_{10}$ contains fewer items than \textsc{MMLU-Pro}$_4$ because of the 500-token length limit: with 10 options, some questions exceed the limit that they satisfy with 4 options.}
\label{tab:model_performances}
\end{table*}

\clearpage
\section{Hyperparameters}
\label{app:hyperparameters}
Hyperparameters used to train models are shown in Table~\ref{tab:hyperparameters}.
\begin{table}[h]
\centering
\small
\renewcommand{\arraystretch}{1.15}
\begin{tabular}{@{}ll@{}}
\toprule
\textbf{Hyperparameter} & \textbf{Value} \\
\midrule
Optimizer & Paged AdamW 8bits \\
Learning rate & $5 \times 10^{-5}$ \\
LR schedule & Linear \\
Warmup ratio & $0.1$ \\
Weight decay & $10^{-2}$ \\
Batch size & $16$ \\
Train sampling strategy & by length \\
\bottomrule
\end{tabular}
\caption{Hyperparameters used for calibration training. The same values are used for all models and datasets.}
\label{tab:hyperparameters}
\end{table}
\clearpage
\section{Output consistency tracking emerges early in training}
\label{sec:development}

\begin{figure}[h]
    \centering
    \includegraphics[width=1.0\linewidth]{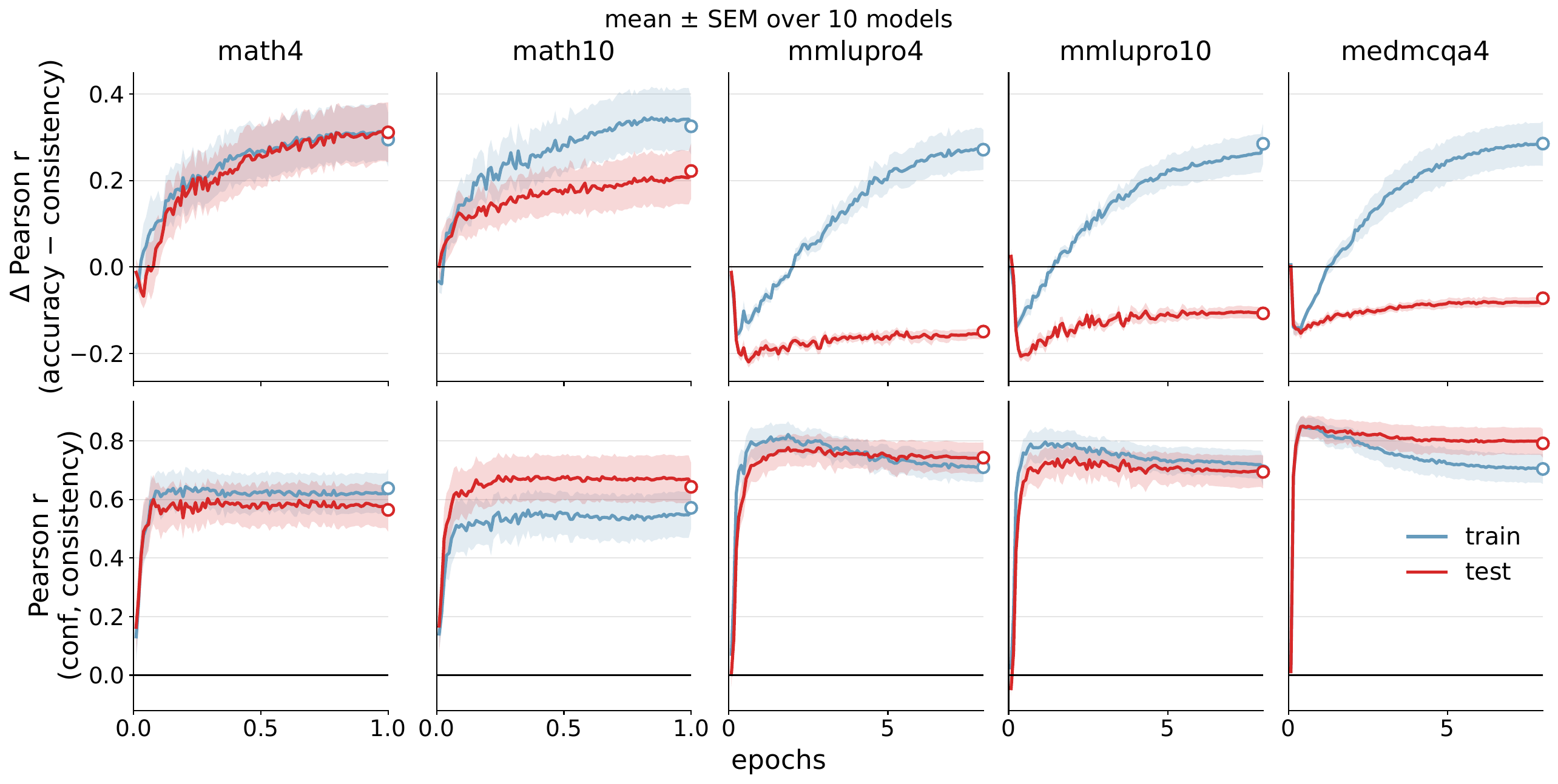}
    \caption{\textbf{Development of output consistency tracking during training.} Each column shows one training dataset. The top row represents $\Delta r$ over training; positive values indicate that confidence tracks true accuracy more than output consistency, negative values the reverse. Bottom row represents Pearson correlation between trained confidence and output consistency. Curves are computed on the train and test splits of the training dataset. Shaded areas: s.e.m.\ across the 10 models. For this specific experiment models were trained on \textsc{MedMCQA} for 8 epochs instead of 1 to investigate the overfitting regime.}
    \label{fig:development}
\end{figure}

We next investigate how the two signals emerge during training. Figure~\ref{fig:development} shows, for each dataset, $\Delta r$ and the correlation between trained confidence and output consistency over training, on both the training split and the test split of the training dataset. On all datasets, the correlation with output consistency peaks early in training and then remains stable. On \textsc{MMLU-Pro} and \textsc{MedMCQA}, confidence initially tracks output consistency more than accuracy on both splits ($\Delta r < 0$). As training proceeds, $\Delta r$ increases on the training split until confidence tracks true accuracy almost perfectly, whereas it remains negative on the test split. Models thus first learn to track output consistency, and only later learn the accuracy of individual training questions, a form of accuracy tracking that does not extend to the test split. On \textsc{Math}, in contrast, $\Delta r$ is negative only very briefly at the start of training, then quickly becomes positive on both splits, earlier than on the other datasets. We attribute this difference to the high similarity between the math questions, accuracy tracking may be learnt fast enough due to their low variance for accuracy tracking to dominate in only a few steps of training. Nevertheless, models trained on \textsc{Math} still track output consistency on the other datasets (Figure~\ref{fig:cross_perf} in Main text). 

\clearpage
\section{Middle Probe}
\label{app:middle_probe}
The confidence matching output consistency results could be linked to the fact that consistency is inherently encoded in the logits, and therefore we could find traces of it in the output of the transformer block which could contaminate the confidence probe trained at the end of the network. To verify whether our observations of output consistency tracking are due to this form of contamination we retrained all networks from scratch to replicate all the previous experiments this time putting the probe at the middle of the network instead of the end. By doubling the rank of LoRA adapters (from 8 to 16) we could conserve the same number of training parameters and kept the same hyperparameters for all experiments.

First, Figure~\ref{fig:trained_perf_mid} in Appendix~\ref{app:individual_plots} shows that LLMs all learnt to predict an accuracy much higher than the baseline verbalised accuracy. Nonetheless, this performance is still on average lower than with the probe at the end of the network ($r_{mid}=0.55\pm 0.04$ against $r_{end}=0.73\pm 0.03$ — s.e.m. computed on models variance).

\begin{figure}[h]
    \centering
    \includegraphics[width=1.0\linewidth]{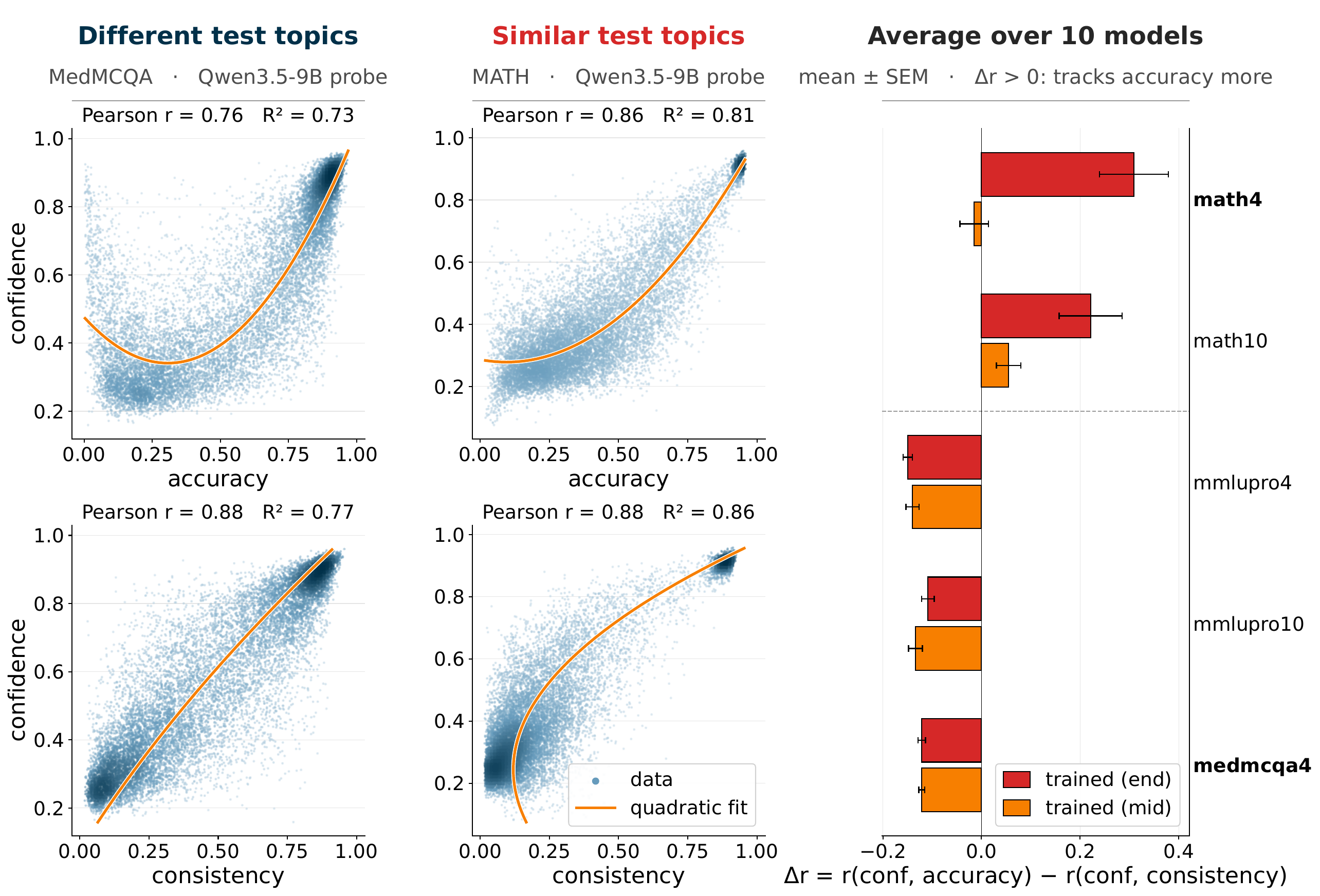}
    \caption{\textbf{Relationship between predicted accuracy, true accuracy and output consistency with middle probe.} Left panel shows the trained confidence of Qwen3.5-9B against true accuracy (top) and output consistency (bottom) on the test split of \textsc{MedMCQA}$_4$ and \textsc{Math}$_4$ with the probe located at the middle of the network. Orange lines are quadratic fits ($R^2$ shown above each plot). Right panel represents $\Delta r$ on each test split for both middle and end probes; positive values (red) indicate that confidence tracks true accuracy more than output consistency, negative values (blue) the reverse. Error bars: s.e.m.\ across the 10 models (each averaged over 5 runs).}
    \label{fig:accuracy_consistency_mid}
\end{figure}

Then, looking at the true accuracy vs. trained confidence (mid probe) plots for Qwen3.5-9B (Figure~\ref{fig:accuracy_consistency_mid} left column) we still noticed a very similar U-shape pattern on \textsc{MedMCQA}$_4$ with a better linear relationship between trained confidence and output consistency ($r=0.88$) than with true accuracy ($r=0.76$). Therefore the output consistency tracking pattern remains. However, the other pattern, true accuracy matching greatly diminished ($r=0.86$ for true accuracy tracking and $r=0.88$ for consistency tracking) which is the opposite of what the leak theory would have predicted. Overall on \textsc{MATH} datasets models do not track true accuracy much better than output consistency ($\Delta r=0.02\pm 0.02$) but still predict a confidence closer to output consistency ($\Delta r=-0.13\pm 0.01$ s.e.m. across models). Additionally, the output consistency prediction over true accuracy prediction with the middle probe is quantitatively comparable with the end probe ($\Delta r=-0.13\pm 0.01$ for middle and $\Delta r=-0.12\pm 0.01$ for end probe).

\begin{figure}[h]
    \centering
    \includegraphics[width=1.0\linewidth]{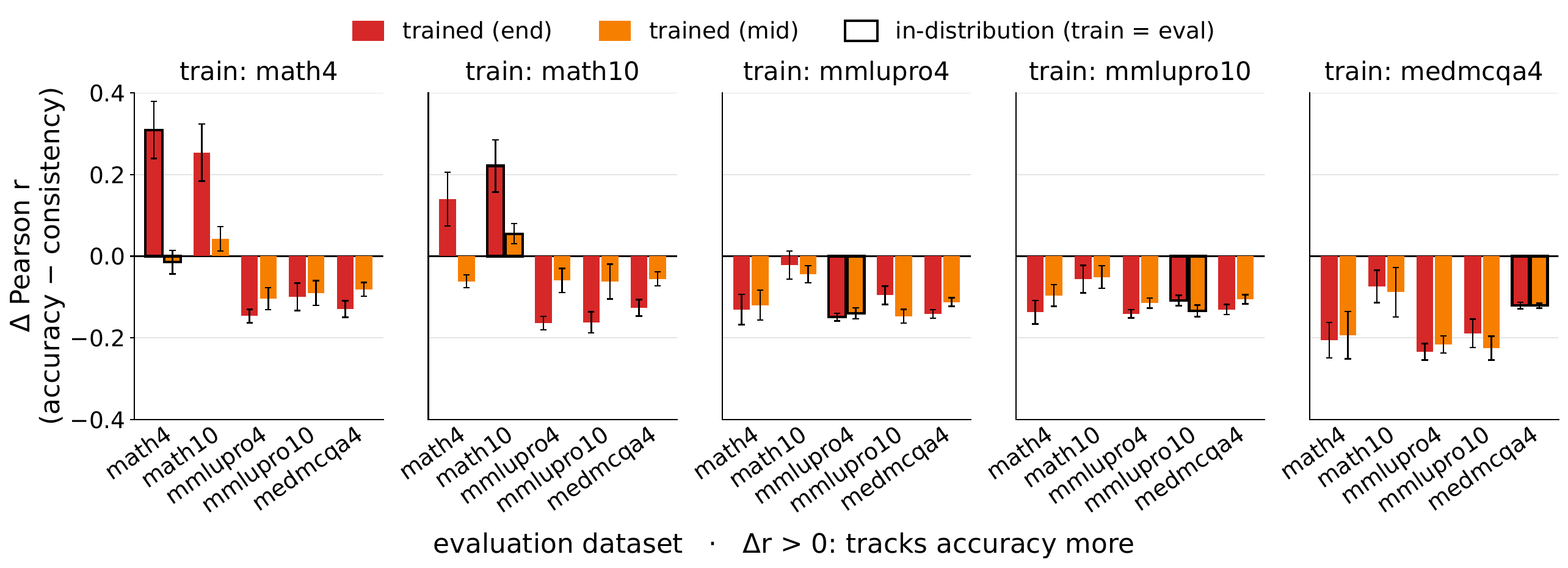}
    \caption{\textbf{Cross-dataset evaluation of trained confidence with the middle probe.} Each panel shows models trained on one dataset and evaluated on the test split of all five datasets. Bars represent $\Delta r$; positive values (red) indicate that confidence tracks true accuracy more than output consistency, negative values (blue) the reverse. In-distribution evaluations (tested on the test split from the training dataset - thematically distinct but closer with respect to the other datasets) are highlighted. Error bars: s.e.m.\ across the 10 models (each averaged over 5 runs).}
    \label{fig:cross_perf_mid}
\end{figure}

When testing models trained on a specific dataset on all the others, consistency tracking results remain consistent: true accuracy prediction on \textsc{MATH} datasets falls by a large margin ($\Delta r=0.01\pm 0.02$ s.e.m. across models) but they keep generalising output consistency on the other datasets ($\Delta r=-0.07\pm 0.02$). On the other hand, the models trained on variants of \textsc{MMLU-PRO} and \textsc{MedMCQA} show a similar $\Delta r$ than with the end probe ($\Delta r=-0.13 \pm 0.02$ for middle probe vs. $\Delta r=-0.12\pm 0.02$ for end probe — see Figure~\ref{fig:cross_perf_mid}).

\begin{figure}[h]
    \centering
    \includegraphics[width=1.0\linewidth]{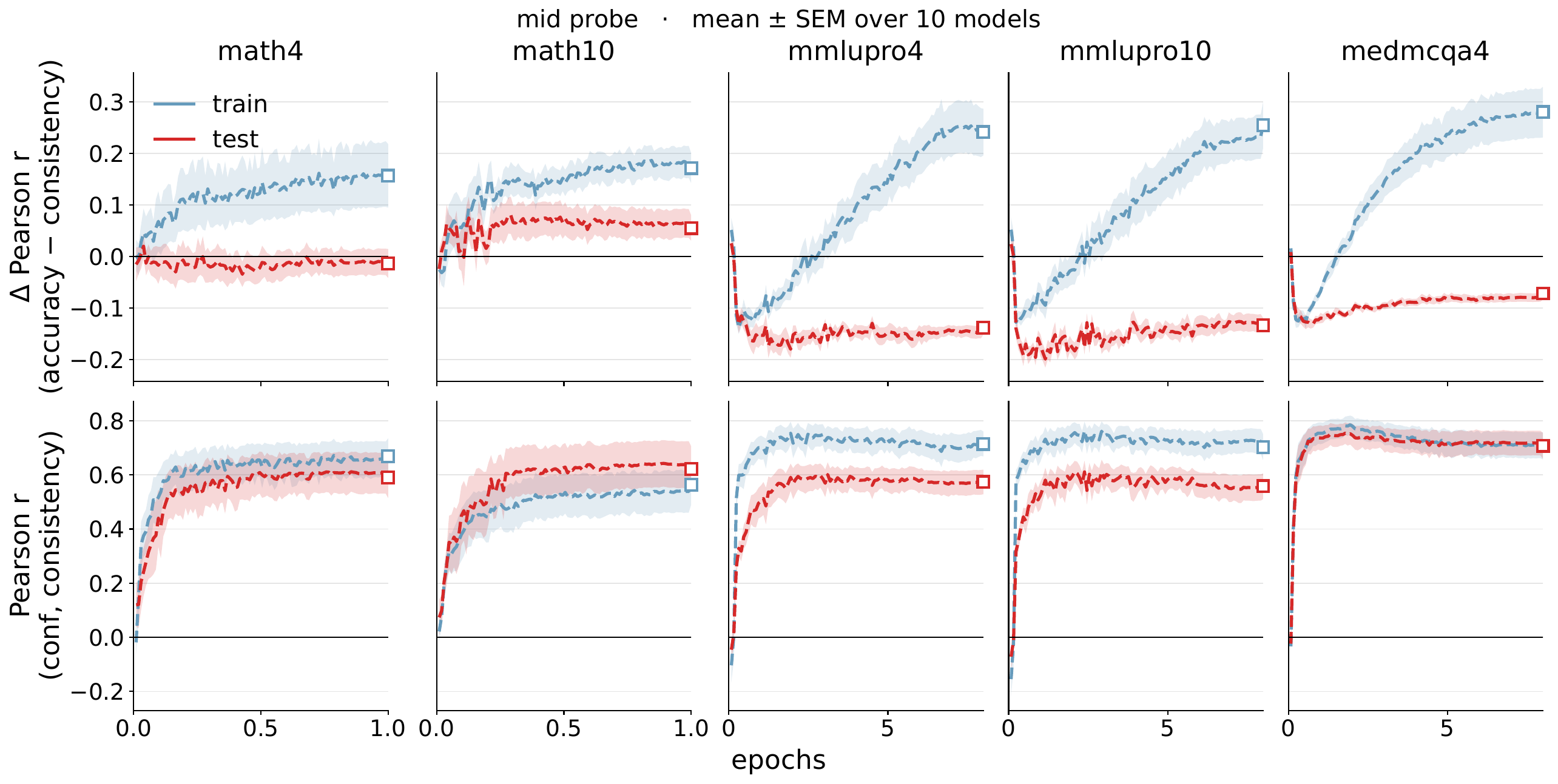}
    \caption{\textbf{Development of output consistency tracking during training with the middle probe.} Each column shows one training dataset. The top row represents $\Delta r$ over training; positive values indicate that confidence tracks true accuracy more than output consistency, negative values the reverse. Bottom row represents Pearson correlation between trained confidence and output consistency. Curves are computed on the train and test splits of the training dataset. Shaded areas: s.e.m.\ across the 10 models.}
    \label{fig:development_mid}
\end{figure}

The acquisition of these capabilities remain very similar than with the end probe: models learn output consistency tracking very early in training in \textsc{MMLU-PRO} and \textsc{MedMCQA} dataset variants but shift faster on \textsc{MATH} datasets yet do not predict true accuracy much better than output consistency (see Figure \ref{fig:development_mid}).

\begin{figure}[h]
    \centering
    \includegraphics[width=1.0\linewidth]{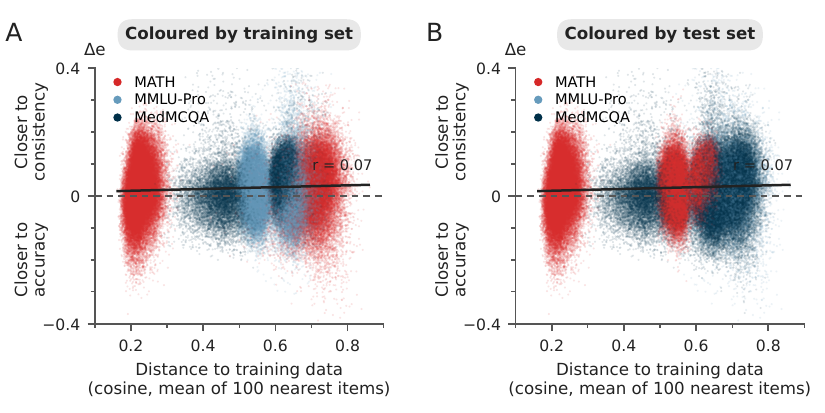}
\caption{\textbf{Distance to the training data and the signal tracked by trained confidence with
the mid probe.} Each point is one test question for one training dataset; its y-value is
$\Delta e$ computed from accuracy, consistency and confidence averaged over 10 models $\times$
5 training runs. Negative values mean confidence is closer to true accuracy, positive values
closer to output consistency. Points are colored by training dataset (A) or test dataset (B).
The x-axis is the mean cosine distance to the 100 nearest training questions. Black line:
linear fit ($r = 0.07$); dashed: $\Delta e = 0$; 0.10\% of points beyond $\pm 0.4$ not shown.}
    \label{fig:distance_mid}
\end{figure}

Lastly, because of this decrease in true accuracy on the \textsc{MATH} variants, in Figure~\ref{fig:distance_mid} distance is no longer a good indicator of $\Delta e$ ($r=0.07$ for mid probe against $r=0.49$ for end probe), the relative prediction error between accuracy over consistency ($\Delta e>0$ means the model predicts consistency better than accuracy and vice versa). Indeed, with the middle probe, confidence seems to generally always be predictive of output consistency above true accuracy.

These results highlight the fact that this output consistency prediction does not arise from a leaking effect from the end of the transformer network but is far more ingrained in the LLM itself. Moreover, when using the middle probe, this time, it is the true accuracy tracking behavior that disappeared from our experiments suggesting this other form of metacognition could emerge from the latter parts of the network. This paper does not investigate in details where both of these metacognitive signals emerge from but suggest output consistency information arises early in training while this second form of metacognition may requires specific information only available in the later parts of the network.

\clearpage
\section{Training curves}
\label{app:training_curves}
\begin{figure}[h]
    \centering
    \includegraphics[width=1.0\linewidth]{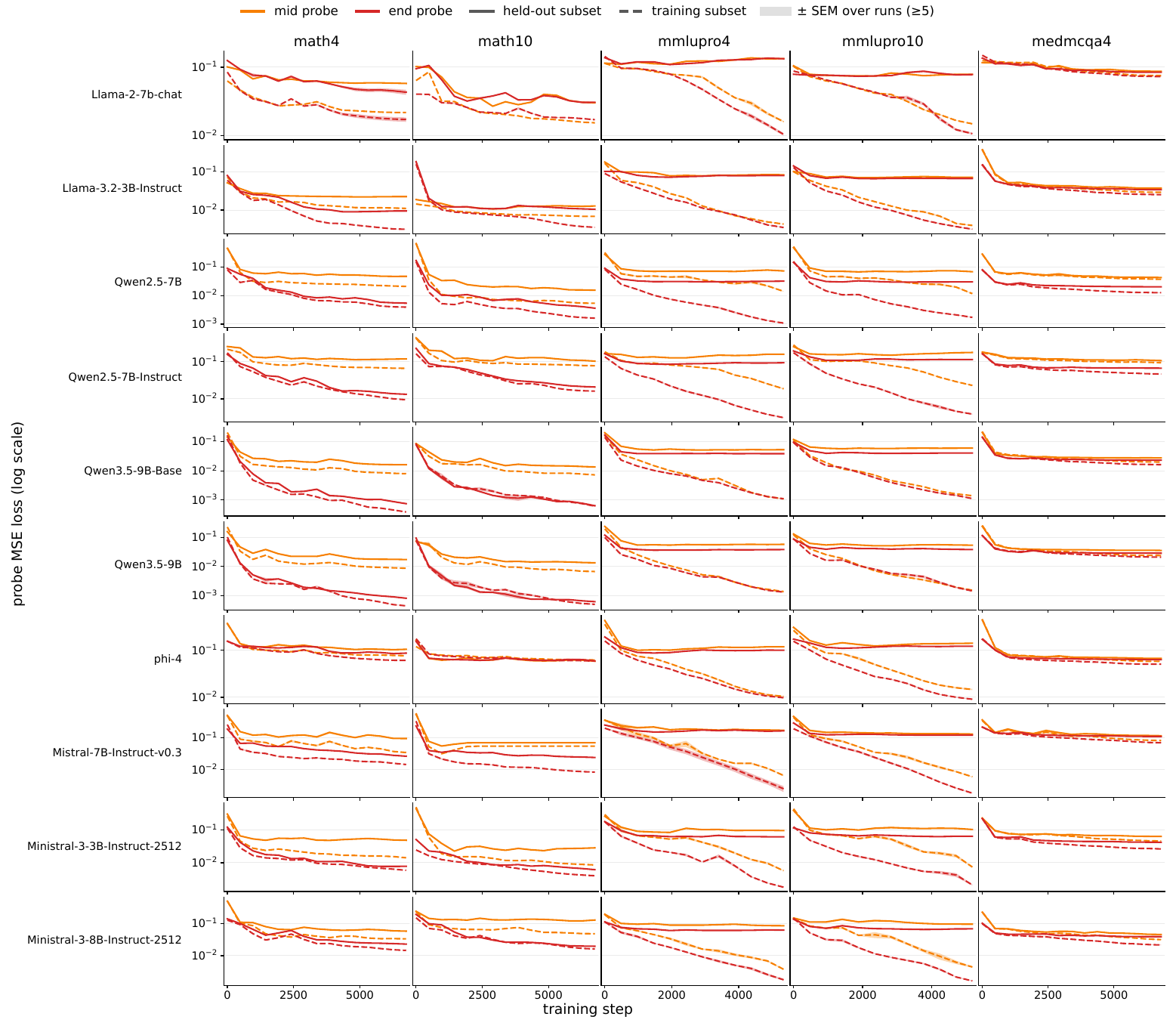}
    \caption{\textbf{Loss during training for both end and mid probes, train and test sets.}}
    \label{fig:loss}
\end{figure}

This section presents the training loss curves in Figure~\ref{fig:loss}.

\clearpage
\section{Individual model plots}
\label{app:individual_plots}
\subsection{Training plots}
\begin{figure}[h]
    \centering
    \includegraphics[width=1.0\linewidth]{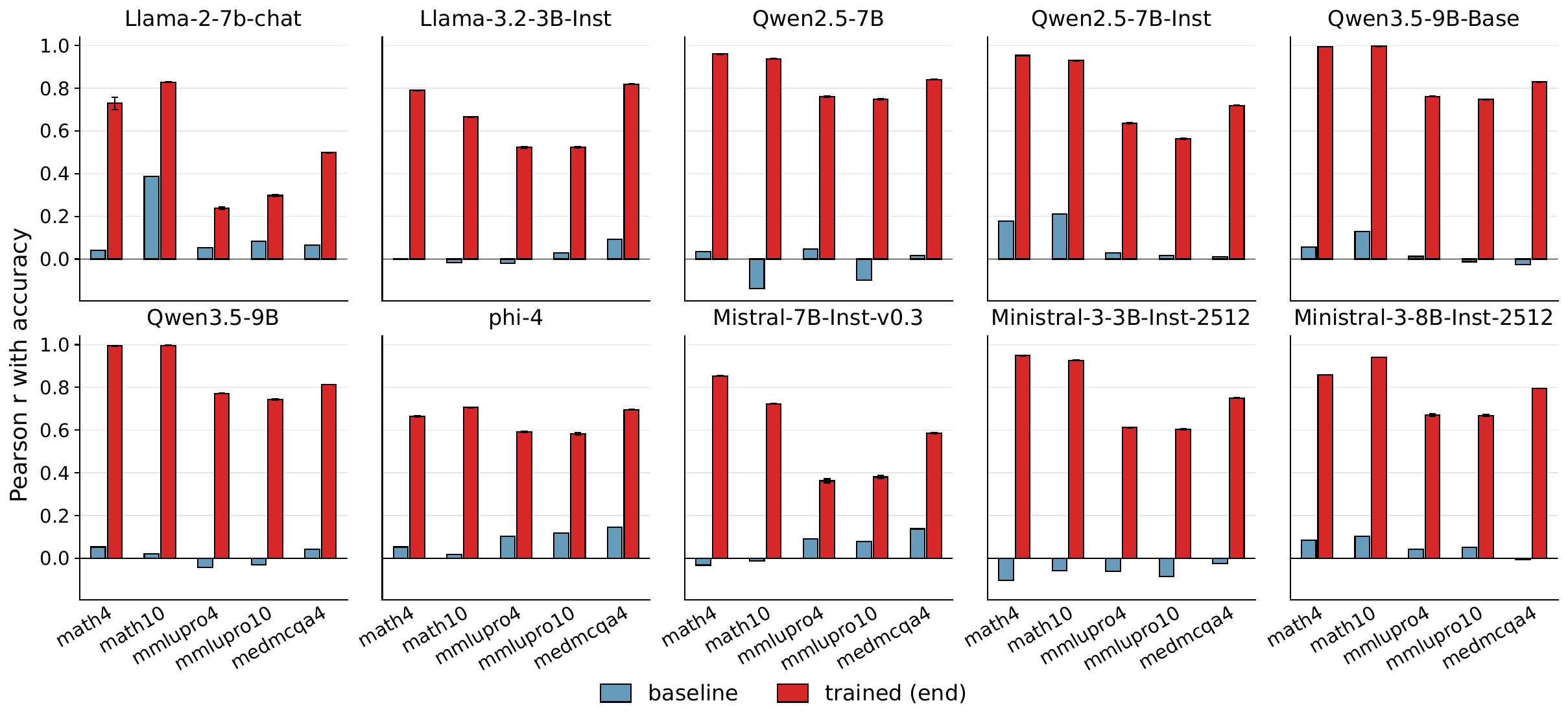}
    \caption{\textbf{Confidence performances before and after training on test splits for each dataset.} Bars represent the pearson r between accuracy and verbalised predicted accuracy (blue) or trained predicted accuracy (red) on each of the 5 datasets and the 10 models. Each bar is the average of 5 runs on the same dataset x model and errorbars represent the stderr between these trainings.}
    \label{fig:trained_perf}
\end{figure}

\begin{figure}[h]
    \centering
    \includegraphics[width=1.0\linewidth]{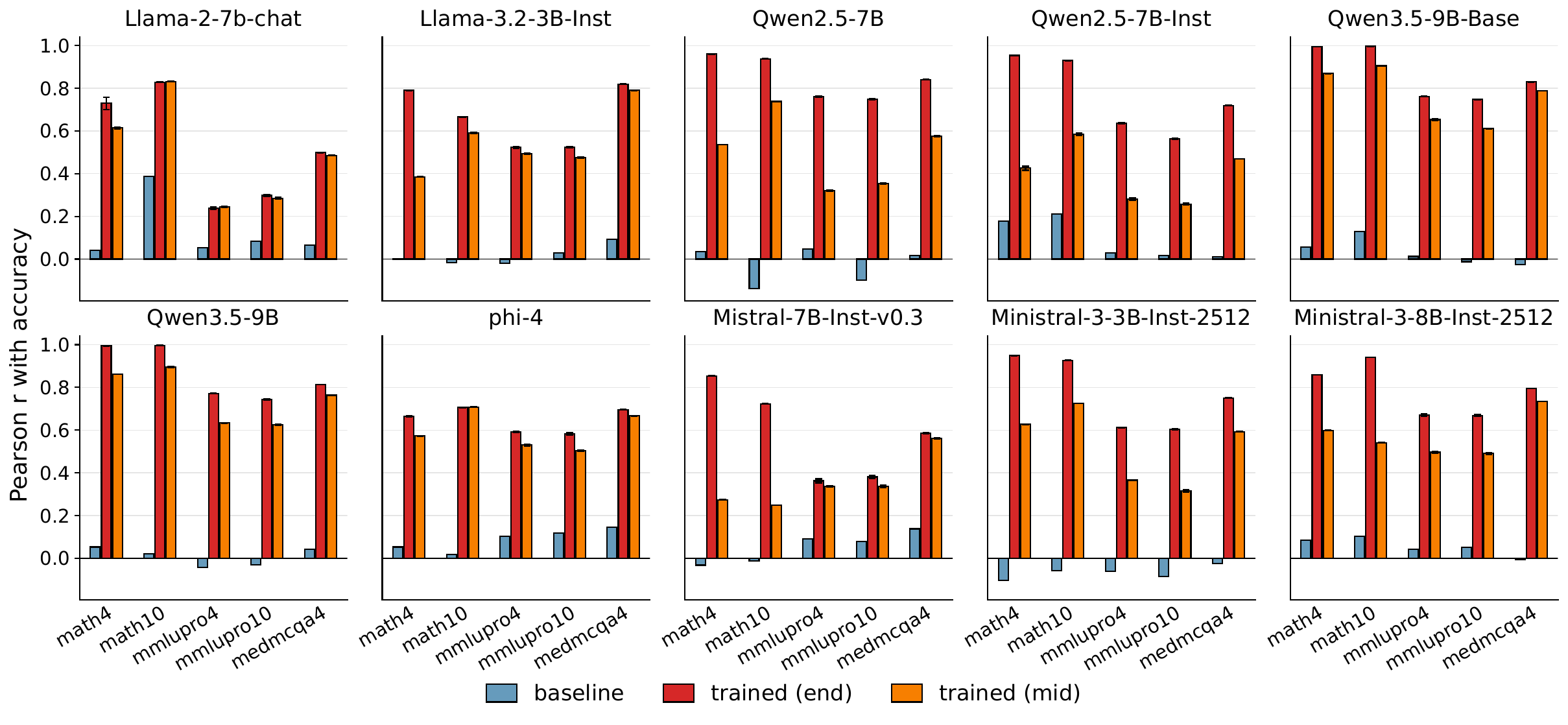}
    \caption{\textbf{Confidence performances before and after training on test splits for each dataset on both probes.} Bars represent the pearson r between accuracy and verbalised predicted accuracy (blue), trained predicted accuracy on the end probe (red) or trained predicted accuracy on the mid probe (orange) on each of the 5 datasets and the 10 models. Each bar is the average of 5 runs on the same dataset x model and errorbars represent the stderr between these trainings.}
    \label{fig:trained_perf_mid}
\end{figure}

\begin{figure}[h]
    \centering
    \includegraphics[width=1.0\linewidth]{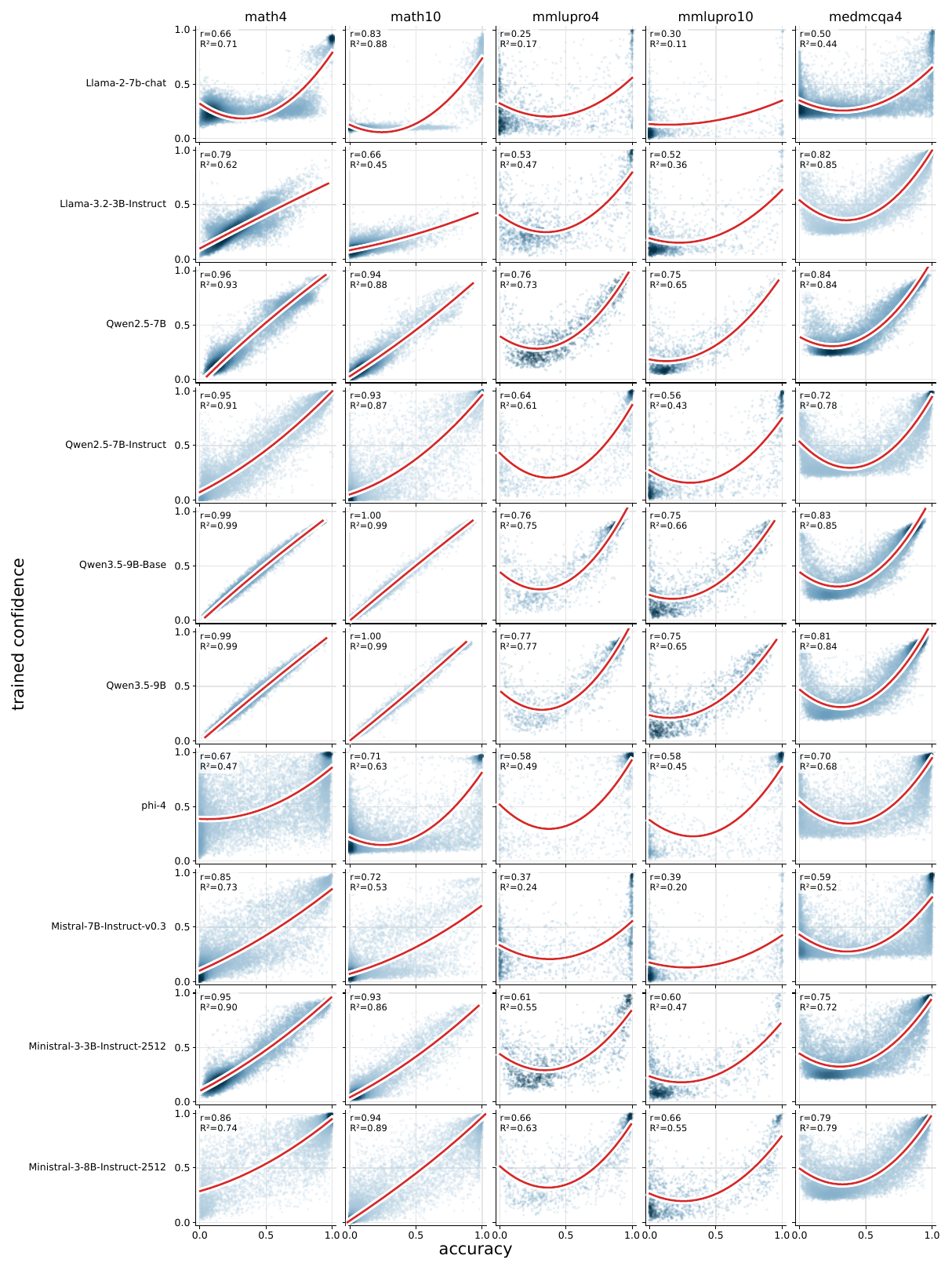}
    \caption{\textbf{Trained confidence vs. accuracy for all models and datasets with the end probe.} X axis is the datasets, Y axis the model. Line is a 2nd order polynomial fitted on the data, $R^2$ its fit coefficient and $r$ is the Pearson correlation coefficient.}
    \label{fig:accuracy_scatter_end}
\end{figure}

\begin{figure}[h]
    \centering
    \includegraphics[width=1.0\linewidth]{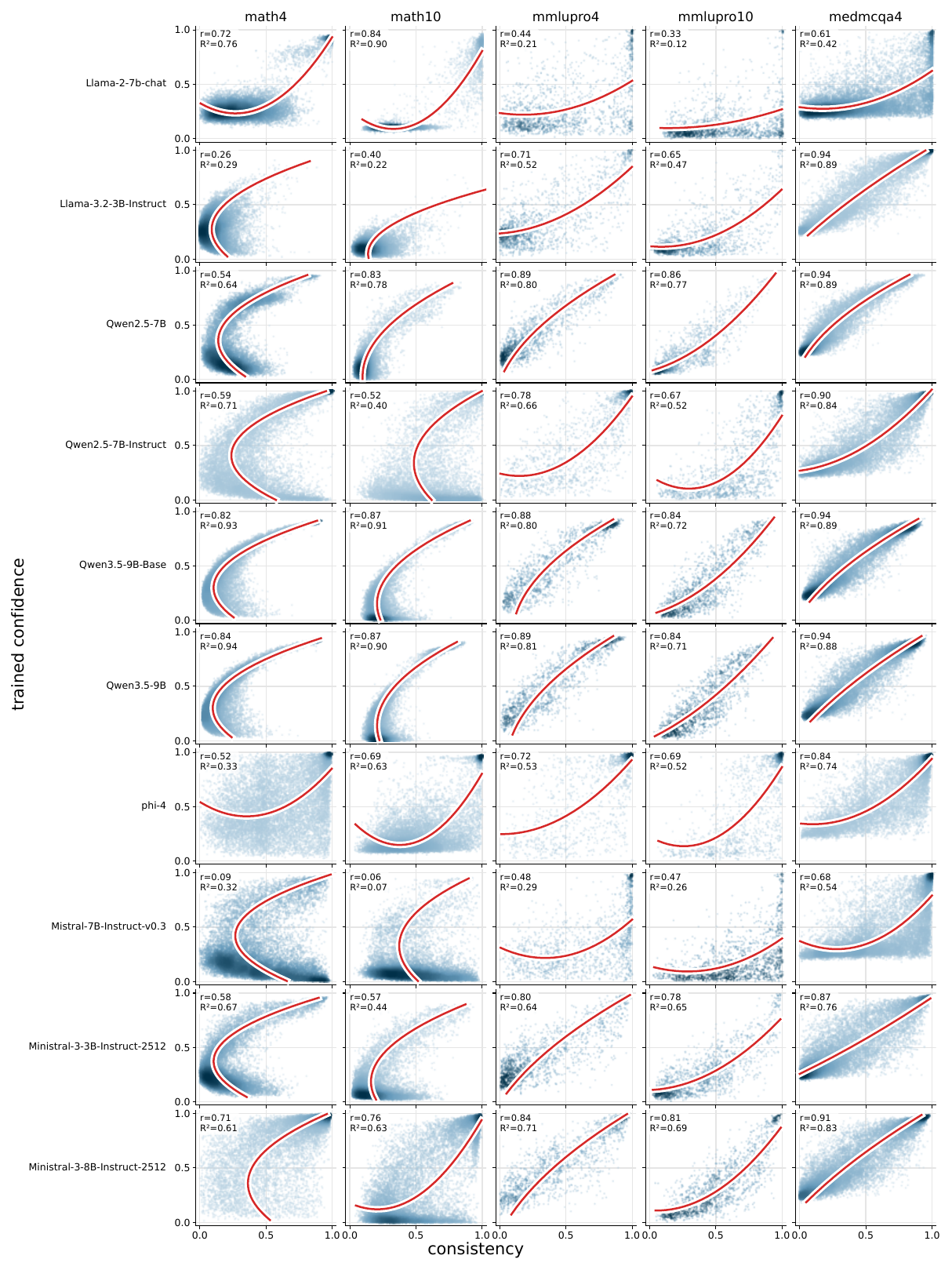}
    \caption{\textbf{Trained confidence vs. consistency for all models and datasets with the end probe.} X axis is the datasets, Y axis the model. Line is a 2nd order polynomial fitted on the data, $R^2$ its fit coefficient and $r$ is the Pearson correlation coefficient.}
    \label{fig:consistency_scatter_end}
\end{figure}

\begin{figure}[h]
    \centering
    \includegraphics[width=1.0\linewidth]{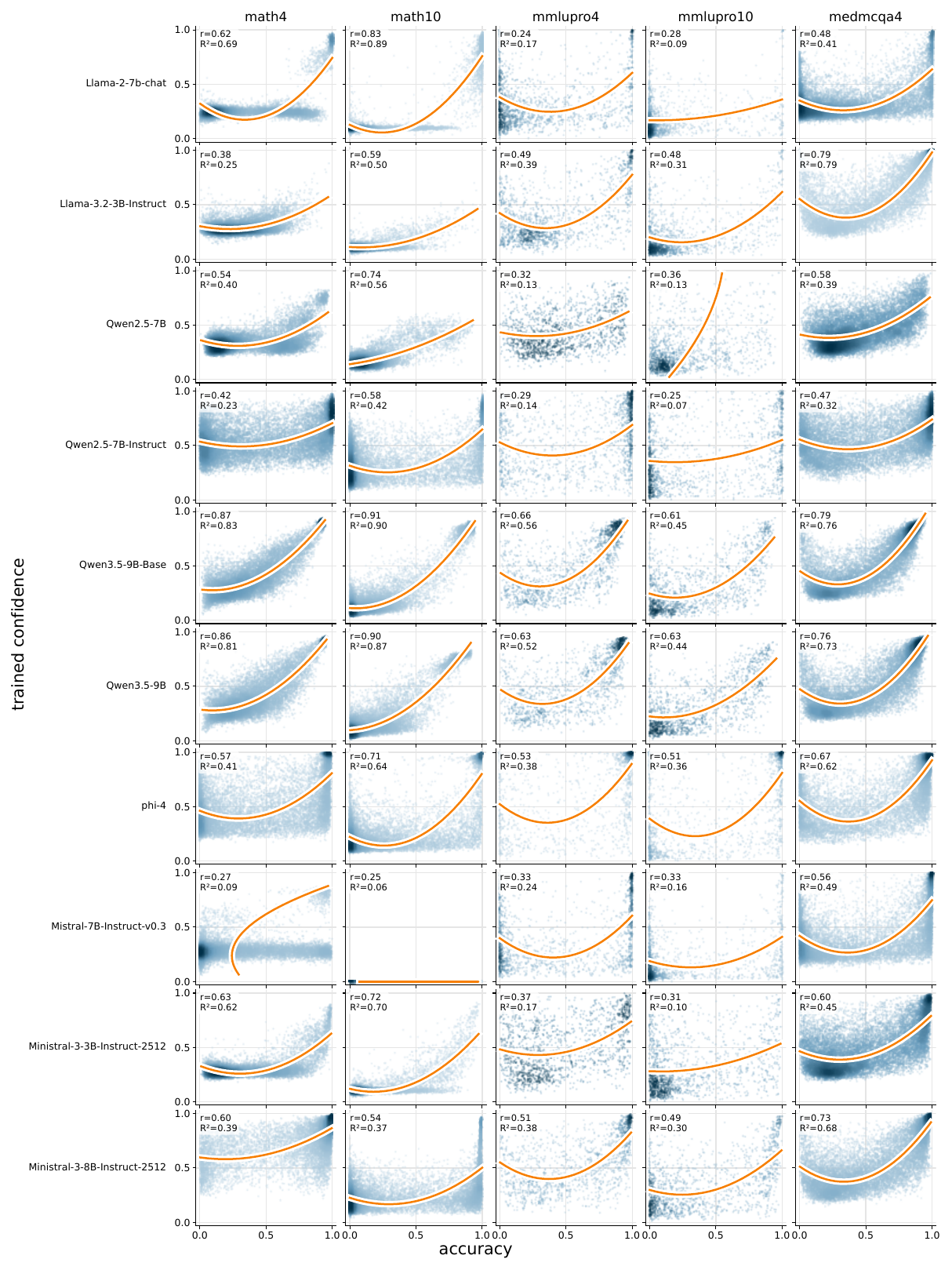}
    \caption{\textbf{Trained confidence vs. accuracy for all models and datasets with the mid probe.} X axis is the datasets, Y axis the model. Line is a 2nd order polynomial fitted on the data, $R^2$ its fit coefficient and $r$ is the Pearson correlation coefficient.}
    \label{fig:accuracy_scatter_mid}
\end{figure}

\begin{figure}[h]
    \centering
    \includegraphics[width=1.0\linewidth]{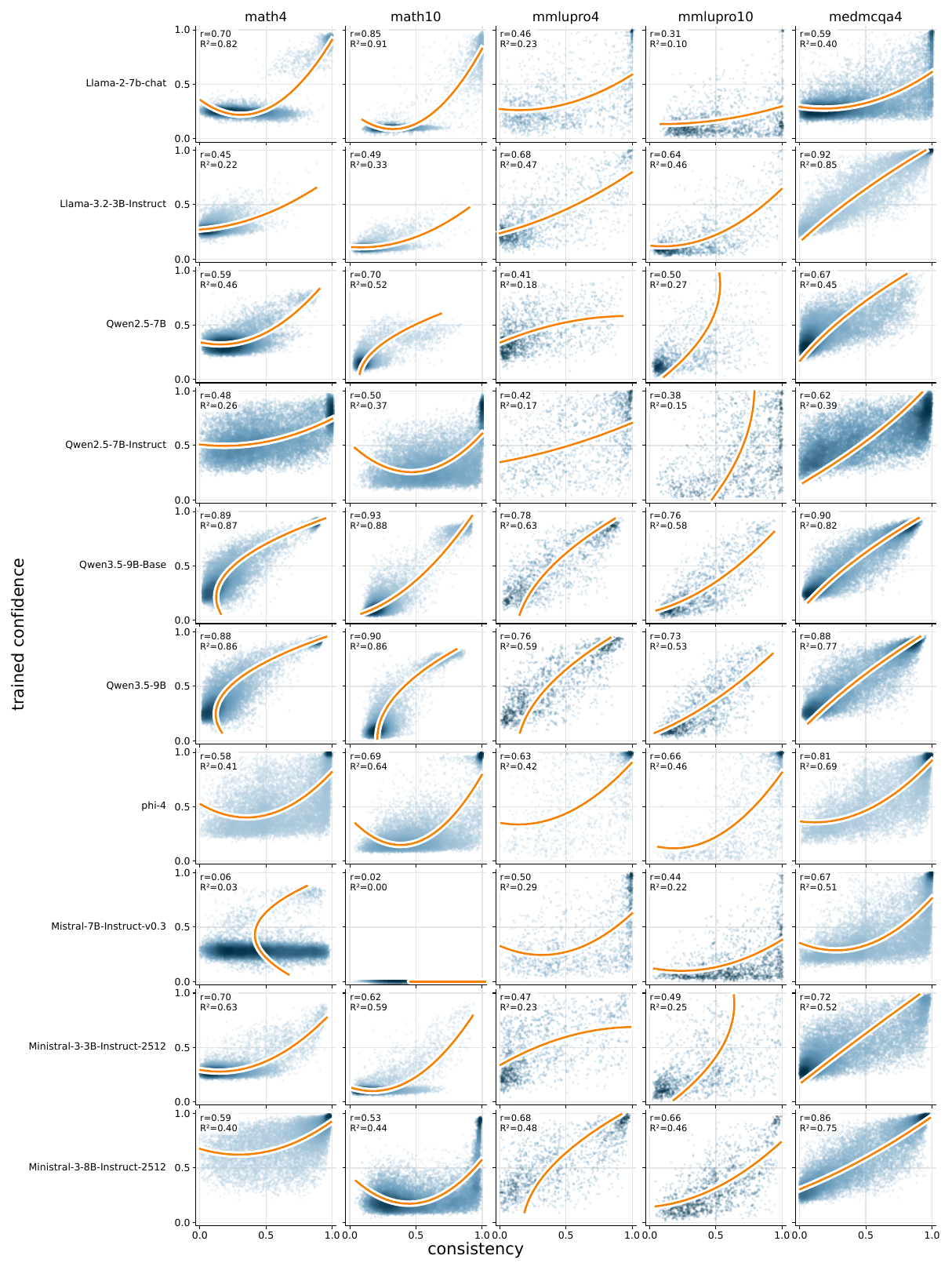}
    \caption{\textbf{Trained confidence vs. consistency for all models and datasets with the mid probe.} X axis is the datasets, Y axis the model. Line is a 2nd order polynomial fitted on the data, $R^2$ its fit coefficient and $r$ is the Pearson correlation coefficient.}
    \label{fig:consistency_scatter_mid}
\end{figure}

\begin{figure}[h]
    \centering
    \includegraphics[width=1.0\linewidth]{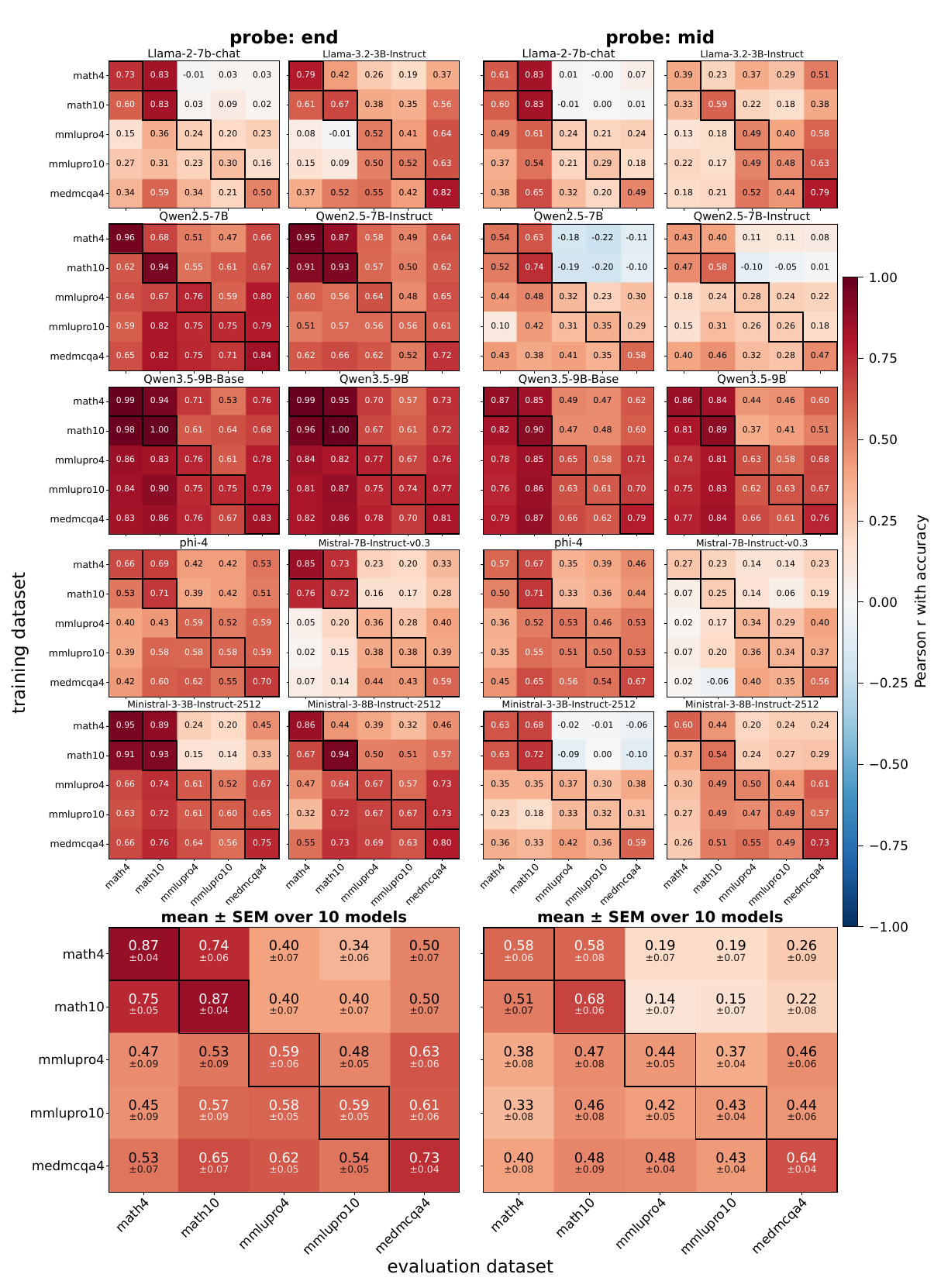}
    \caption{\textbf{Cross dataset true accuracy correlation with trained confidence across both probes.} Left column represents end probe results, right column, mid probe results and bottom plots show average + s.e.m. on all models together.}
    \label{fig:accuracy_correlation}
\end{figure}

\begin{figure}[h]
    \centering
    \includegraphics[width=1.0\linewidth]{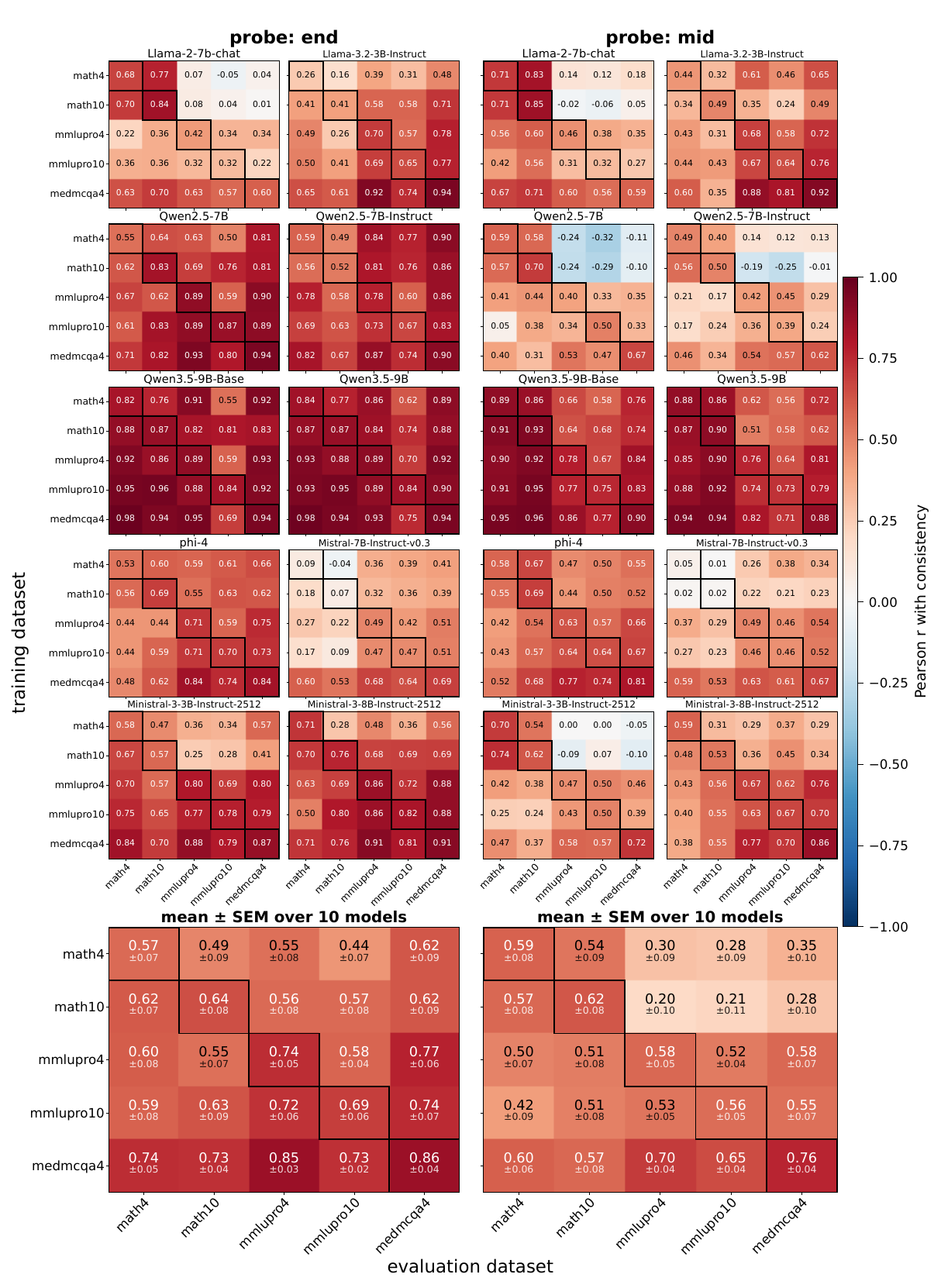}
    \caption{\textbf{Cross dataset output consistency correlation with trained confidence across both probes.} Left column represents end probe results, right column, mid probe results and bottom plots show average + s.e.m. on all models together.}
    \label{fig:consistency_correlation}
\end{figure}

\begin{figure}[h]
    \centering
    \includegraphics[width=1.0\linewidth]{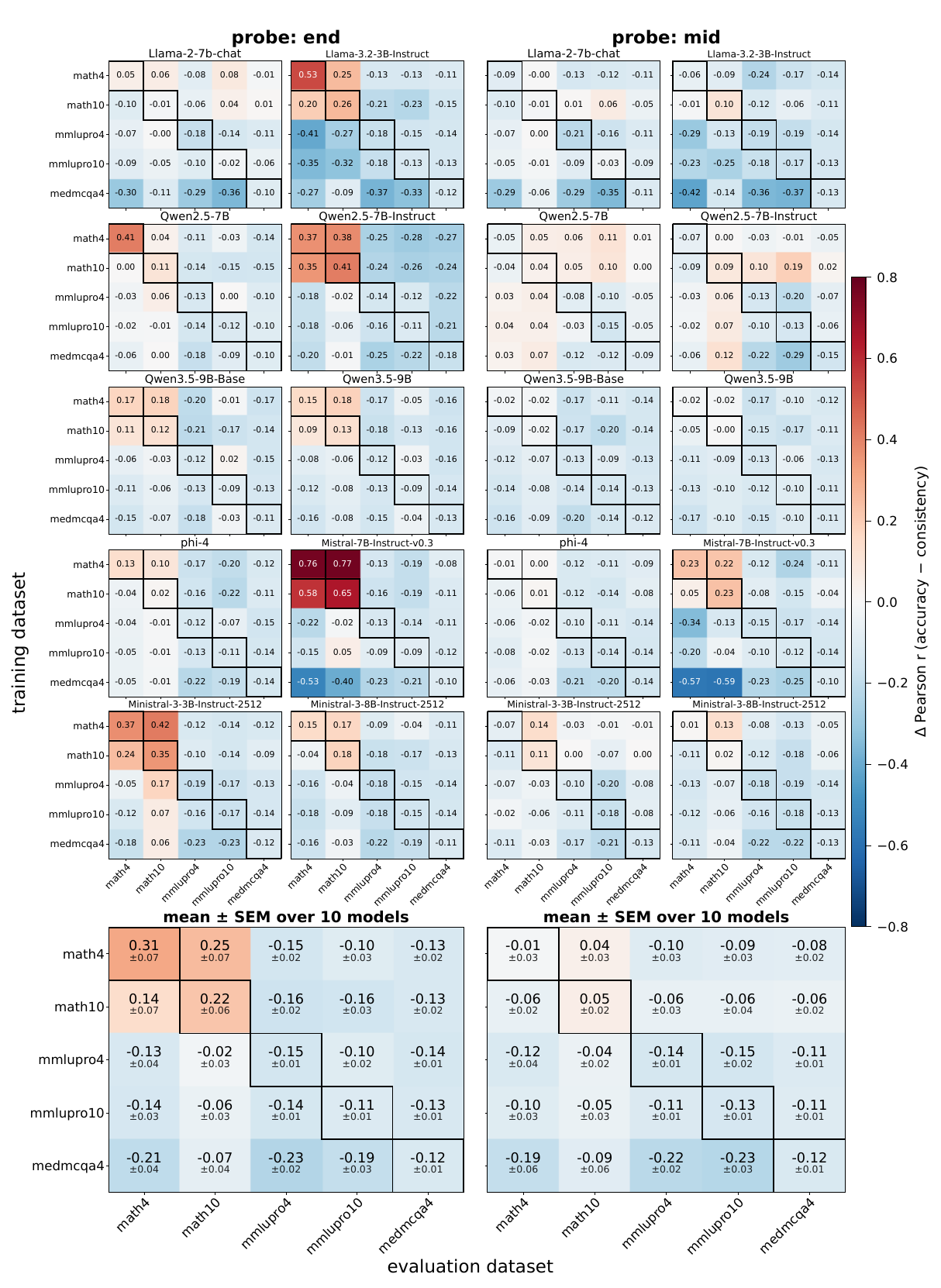}
    \caption{\textbf{Cross dataset $\Delta r$ across both probes.} Left column represents end probe results, right column, mid probe results and bottom plots show average + s.e.m. on all models together.}
    \label{fig:deltar}
\end{figure}

This section contains figures showing individual models behaviors. Figure~\ref{fig:trained_perf} shows correlations between verbalised confidence and trained confidence (end probe) with true accuracy. Figure~\ref{fig:trained_perf_mid} also include the middle probe alongside the others. Figures~\ref{fig:accuracy_scatter_end} \ref{fig:consistency_scatter_end} \ref{fig:accuracy_scatter_mid} \ref{fig:consistency_scatter_mid} show respectively the scatter plot for trained confidence vs. accuracy or consistency for both end and mid probes. Figure~\ref{fig:accuracy_correlation} represents all the correlations between true accuracy and predicted accuracy (left column is end probe, right is mid probe) for models trained on the train set of a datasets and tested on all the other datasets' test set. Figure~\ref{fig:consistency_correlation} shows the same type of representations but for correlation between output consistency and predicted accuracy and finally Figure~\ref{fig:deltar} represents $\Delta r$ for each individual model in the same settings.

\end{document}